\documentclass[10pt,twocolumn,letterpaper]{article}

\usepackage{iccv}
\usepackage{times}
\usepackage{epsfig}
\usepackage{graphicx}
\usepackage{amsmath}
\usepackage{amssymb}
\usepackage{booktabs}
\usepackage{tabularx}
\usepackage{multirow}

\usepackage{float}

\usepackage{enumitem}

\usepackage[dvipsnames,table]{xcolor}
\usepackage[pagebackref=true,breaklinks=true,letterpaper=true,colorlinks,bookmarks=false,citecolor=Orange]{hyperref}
\usepackage[accsupp]{axessibility}  %

\iccvfinalcopy %

\def\iccvPaperID{6622} %

\ificcvfinal\fi

\newcolumntype{F}{>{\centering\arraybackslash}X}

\newif\ifsubmit
\definecolor{author_colorA}{rgb}{0,0.5,1}
\definecolor{author_colorB}{rgb}{0.2,.64,0}
\definecolor{author_colorC}{rgb}{1,0,1}
\definecolor{author_colorD}{rgb}{0,.54,.31}
\definecolor{author_colorE}{rgb}{1.,.24,.51}
\definecolor{changes_color}{rgb}{0.05,0.5,0.3}
\definecolor{mathbrace_color}{rgb}{0.2,0.5,1.0}
\iftrue

\submittrue

\ifsubmit
    \newcommand{\abe}[1]{}
    \newcommand{\authorB}[1]{}
    \newcommand{\authorC}[1]{}
    \newcommand{\ruyu}[1]{}
    \newcommand{\sid}[1]{}
    \newenvironment{changes}
      {
      }
      {}
\else
    \newcommand{\abe}[1]{\textbf{\textcolor{blue}{AD: #1}}}
    \newcommand{\authorB}[1]{\textbf{\textcolor{author_colorB}{CS: #1}}}
    \newcommand{\authorC}[1]{\textsf{\textcolor{author_colorC}{[{\bf AUTHORB}: #1]}}}
    \newcommand{\ruyu}[1]{\textcolor{author_colorD}{[{\bf RY}: #1]}}

    \newcommand{\sid}[1]{\textcolor{author_colorC}{[{\bf Sid}: #1]}}

\fi

\usepackage{amsmath,amsfonts,bm}

\def\eqref#1{equation~\ref{#1}}

\def\1{\bm{1}}

\def\rvc{{\mathbf{c}}}
\def\rvd{{\mathbf{d}}}

\def\rvo{{\mathbf{o}}}

\def\rvr{{\mathbf{r}}}

\def\rvt{{\mathbf{t}}}

\def\rvw{{\mathbf{w}}}

\def\rvz{{\mathbf{z}}}

\def\rmE{{\mathbf{E}}}

\def\rmI{{\mathbf{I}}}

\def\rmK{{\mathbf{K}}}

\def\rmR{{\mathbf{R}}}

\DeclareMathAlphabet{\mathsfit}{\encodingdefault}{\sfdefault}{m}{sl}
\SetMathAlphabet{\mathsfit}{bold}{\encodingdefault}{\sfdefault}{bx}{n}

\begin{document}

\title{Ray Conditioning: Trading Photo-consistency for Photo-realism in Multi-view Image Generation}

\author{Eric Ming Chen$^1$ \qquad Sidhanth Holalkere$^1$ \qquad Ruyu Yan$^1$ \qquad Kai Zhang$^2$ \qquad Abe Davis$^1$\vspace{5pt}\\
$^1$Cornell University \qquad $^2$Adobe Research\vspace{5pt}\\
\url{https://ray-cond.github.io}
}

\maketitle
\ificcvfinal\thispagestyle{empty}\fi

\begin{abstract}
Multi-view image generation attracts particular attention these days due to its promising 3D-related applications, e.g., image viewpoint editing. Most existing methods follow a paradigm where a 3D representation is first synthesized, and then rendered into 2D images to ensure photo-consistency across viewpoints. However, such explicit bias for photo-consistency sacrifices photo-realism, causing geometry artifacts and loss of fine-scale details when these methods are applied to edit real images. To address this issue, we propose ray conditioning, a geometry-free alternative that relaxes the photo-consistency constraint. Our method generates multi-view images by conditioning a 2D GAN on a light field prior. With explicit viewpoint control, state-of-the-art photo-realism and identity consistency, our method is particularly suited for the viewpoint editing task. 
\end{abstract}

\section{Introduction}

    Modeling the distributions of natural images has long been an important problem that is extensively studied. Generative adversarial networks (GANs) and diffusion models are two types of generative models that have successfully shown impressive capabilities of learning image distributions---the generated samples are almost indistinguishable from real photos~\cite{Goodfellow2014GenerativeAN,SohlDickstein2015DeepUL,ho2020denoising,song2021scorebased}.
    
    While optimizing for the photo-realism of individual samples, these generative models rarely allow for multi-view image generation, where photo-consistency matters. Recently, many multi-view image synthesizers have been proposed that try to optimize for both photo-realism and photo-consistency~\cite{Chan2021,zhao-gmpi2022,gu2021stylenerf,orel2022stylesdf}. They generally follow a ``synthesize-3D-then-render" paradigm: 3D representations are synthesized, and then images are rendered (at specified camera poses). Such 3D-aware generative models, especially EG3D~\cite{Chan2021}, achieve high-quality multi-view image generation results, despite being trained only on single-view image data with poses. 

    \begin{figure}[t]
\raggedleft
  \setlength{\tabcolsep}{1.0pt}
  \scalebox{0.88}{
  \begin{tabular}{cl}
    \rotatebox{90}{\hspace{-2.0pt} Pose Cond. \cite{Chan2021}} &  \includegraphics[width=1.0\linewidth]{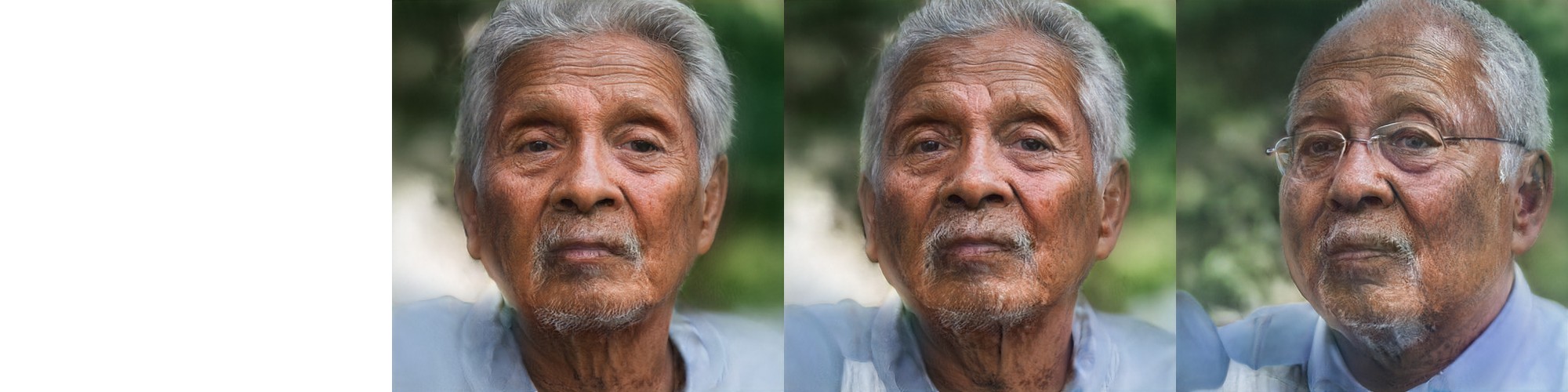}   \\
    \rotatebox{90}{\hspace{8.0pt} EG3D \cite{Chan2021}}              &  \includegraphics[width=1.0\linewidth]{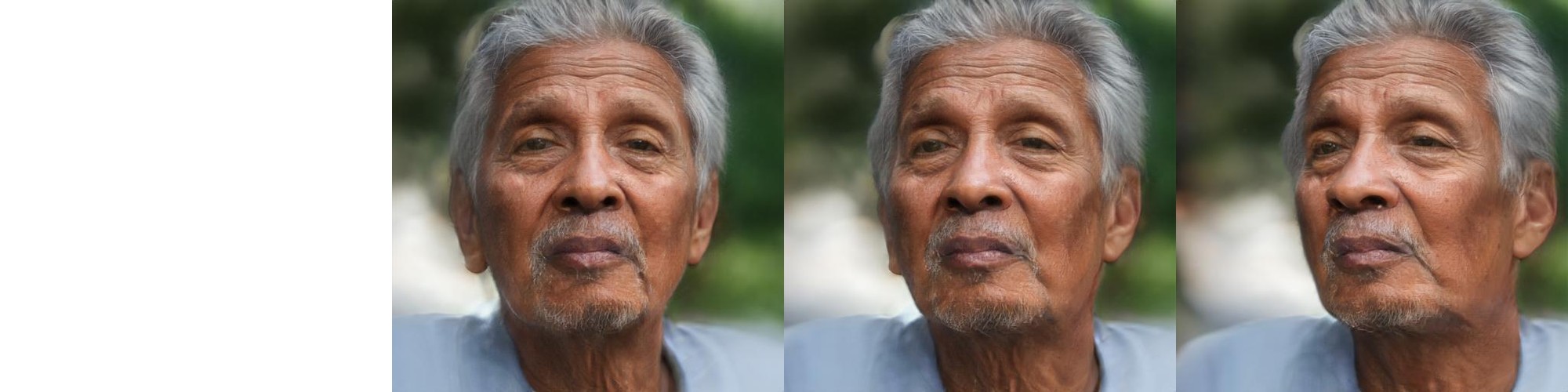}   \\
    \rotatebox{90}{\hspace{6.0pt} Ray Cond.}          &  \includegraphics[width=1.0\linewidth]{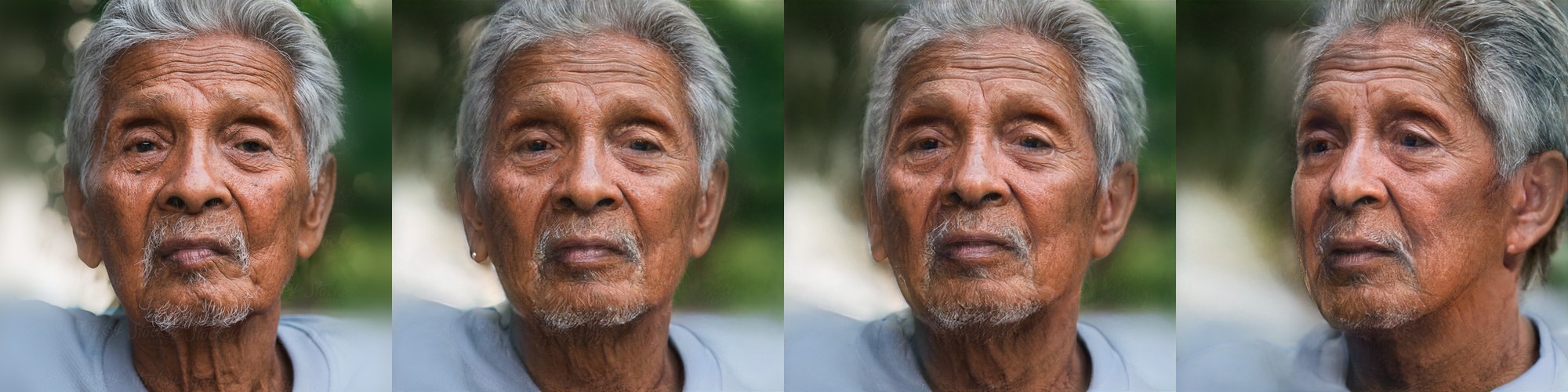} \\
    & \hspace{16.0pt} Input \hspace{26.0pt} Inversion \hspace{48.0pt} Rotation
  \end{tabular}
  }
  \caption{\textbf{Challenges With Viewpoint Editing.} Conditioning a 2D GAN's latent space on pose does not ensure that an identity remains consistent across views. 3D-aware GANs such as EG3D~\cite{Chan2021} struggle to reconstruct high-frequency details such as wrinkles and hair. Our ray conditioning method most faithfully reproduces the input image, and preserves identity when editing the viewpoint. }
  \label{fig:inversion-small}
\end{figure}
    \begin{figure*}[h!]
  \centering
  \includegraphics[width=0.96\linewidth]{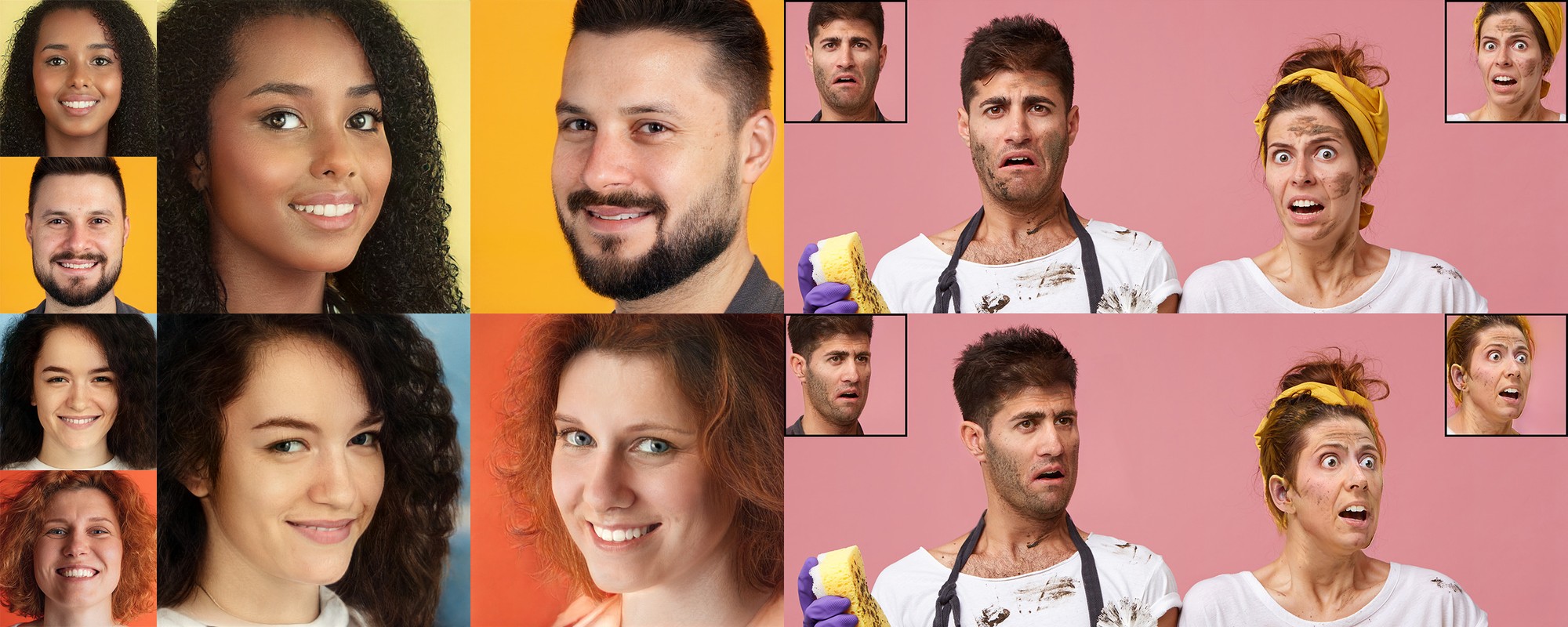}  

  Input \hspace{80pt} Novel Views \hspace{145pt} Photoshop Blending \hspace{60pt}
  
  \caption{\textbf{Viewpoint Editing with Ray Conditioning.} Ray conditioning enables photo-realistic multi-view image editing on natural photos via GAN inversion. 
  The left half shows headshots of four individuals and their corresponding synthesized results from another viewpoint. The right half shows a portrait of two individuals (top row), the GAN inversion results of their faces (top row corners), and the resulting image (bottom row), in which their faces are replaced with synthesized faces looking in a different direction (bottom row corners). To produce the latter, we used Photoshop to blend the synthesized faces with the original image.
  }
  \label{fig:teaser}
\end{figure*}

    However, photo-realism and photo-consistency are oftentimes two competing goals: photo-realism very explicitly favors image quality over control, while photo-consistency more implicitly favors control over quality. A few examples of this conflict include fine-scale detail and view-dependent appearance in the multi-view 3D reconstruction and 3D generation problems. Details are either filtered from 3D representations (leaving them smoother and more diffuse than real ones) or misinterpreted as geometric artifacts~\cite{kaizhang2020}. As shown in Figure~\ref{fig:inversion-small}, although EG3D allows explicit camera control and generates photo-consistent images at different viewpoints, it fails to reproduce the subtle details, e.g., the wrinkles and hairs, in the input image. Conditioning a 2D GAN's latent space on camera pose does not ensure that the identity remains consistent across views.  %

 Our work is motivated by the observation that, for certain classes of images with shared canonical structure, e.g. faces,
 it is possible to achieve viewpoint control without optimizing explicitly for 3D structure. The result is a modified 2D GAN that offers precise control over generated viewpoints without sacrificing photo-realism. Furthermore, we are able to train on data that does not contain multiple viewpoints of any single subject, letting us leverage the same diverse and abundant data used for regular GANs. Our method combines the photo-realism of existing GANs with the control offered by geometric models, outperforming related methods in both generation and inversion quality. This makes our method particularly well-suited for viewpoint editing in static images.

     Key to our method is the proposed \textbf{\textit{ray conditioning}} mechanism that enables explicit viewpoint control. The method is simple. 
     Rather than using a 3D model, our method conditions each pixel in a generated image on the ray through it---a technique inspired by the light field \cite{Levoy1996LightFields,Gortler1996Lumigraph}. The spatial priors of ray conditioning enables the image synthesizer to learn multi-view consistency from only single-view image collections and their estimated poses. By choosing a geometry-free approach, we relax the 3D photo-consistency constraints in exchange for increased photo-realism.
     Despite this, our approach still offers competitive identity preservation capability when editing viewpoints. Figure~\ref{fig:teaser} represents the quality and control we can achieve with ray conditioning. Evaluation on both single-view and multi-view data shows that our method is a significant improvement in image quality over Light Field Networks (LFNs), another geometry-free image synthesizer~\cite{Sitzmann2021LFNs}, demonstrating the promising potential of this line of research. 

    In summary, our contributions are as follows. 
\begin{enumerate}[nosep]
    \item We propose a simple yet effective geometry-free generative model named ray conditioning for multi-view image generation, and show that it achieves greater photo-realism than geometry-based baselines while maintaining explicit control of viewpoints.
    \item We demonstrate the advantages of our method in the downstream application of editing real images' viewpoints where photo-realism is favored over photo-consistency. 
    \item Our ray conditioning method is also the first geometry-free \textit{multi-view} image synthesizer that can generate highly realistic images in high resolution ($1024\times1024$) given only single-view posed image collections.
\end{enumerate}

\section{Related Work}
\noindent\textbf{Image Synthesis and Latent Space Editing.} In the past decade, GANs~\cite{Goodfellow2014GenerativeAN} have revolutionized image synthesis. They are trained by optimizing two neural networks, a generator and a discriminator, at the same time. The generator tries to fool the discriminator by generating fake images that look as real as possible. The discriminator tries to distinguish the synthetic images from the real ones. Once training is complete, the generator is able to generate images that are almost indistinguishable from natural images. Among all the proposed GAN architectures, for its high image quality, StyleGAN~\cite{Karras2018ASG,Karras2019AnalyzingAI,Karras2021AliasFreeGA} is perhaps the most widely adopted. We base our method on StyleGAN2, and enable explicit viewpoint control via ray conditioning. 

Several works have found that the latent spaces of StyleGAN are remarkably linear~\cite{Jahanian*2020steerability,shen2020interfacegan,harkonen2020ganspace,Wu2020StyleSpaceAD}, disentangling attributes such as facial expression, hair color, and pose. Pose disentanglement is of particular interest to the domain of this work, because it serves as a proxy for 3D information. Related work have utilized this property of 2D GANs to discover visual correspondences between images~\cite{peebles2022gansupervised}. However, properly harnessing this ability can be non-trivial. In a latent space, editing directions often have to be found ad-hoc for each dataset, and lack a intuitive interpretation~\cite{shen2020interfacegan}. In contrast, our method allows for interpretable and explicit control of viewpoints.

\medskip

\noindent\textbf{Multi-view Image Synthesis.} Our work is closely related to the direction of multi-view image synthesis, where prior work can be roughly classified into two categories: \textit{geometry-based} and \textit{geometry-free}. 

\textit{Geometry-based} multi-view image generation approaches typically adapt 2D GAN-based image synthesizers in a way that a neural 3D representation is first generated, from which 2D images are then rendered using neural rendering. Representative work in this category include EG3D~\cite{Chan2021}, GMPI~\cite{zhao-gmpi2022}, StyleSDF~\cite{orel2022stylesdf}, StyleNeRF~\cite{gu2021stylenerf}, GIRAFFE~\cite{Niemeyer2020GIRAFFE}, $\pi$-GAN~\cite{chanmonteiro2020pi-GAN}, etc. These methods mainly differ from each other in choice of the 3D representation and rendering algorithm. For instance, EG3D adapts the StyleGAN2 generator to predict a feature volume in a compact triplane representation. They then use volume rendering to render a feature map, which is later decoded into a high-resolution image through a convolutional decoder; on the other hand, GMPI generates a multiplane image RGBA representation~\cite{Zhou2018StereoML} and renders it using homography warping and alpha compositing. These \textit{geometry-based} methods have demonstrated impressive multi-view image synthesis quality given only single-view posed data; the generated images are very view-consistent and detailed. However, we observe that the synthesize-3D-then-render approach they adopt indeed trades some photo-realism for photo-consistency, as shown in Figure~\ref{fig:inversion-small}. Moreover, as noted by concurrent work~\cite{xie2022HFGI3D}, the geometry prior in an image synthesizer also increases the difficulty of inverting a real posed image. Both issues sacrifice the methods' performance in viewpoint editing for real posed images. We seek to circumvent these issues by optimizing for photo-realism and easy invertibility.

\begin{figure*}[h!]
  \centering
  \includegraphics[width=0.9\linewidth]{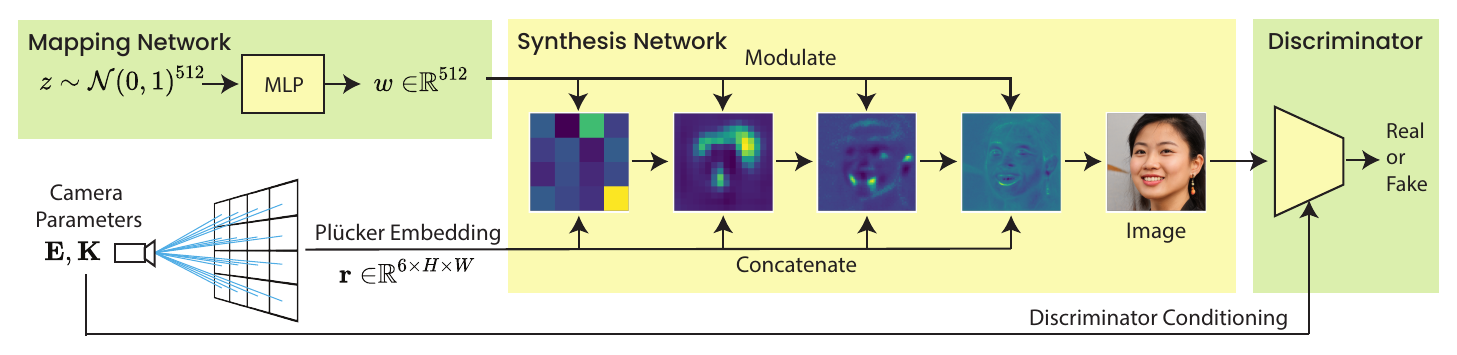}
 \caption{\textbf{The Ray Conditioning Method.} The StyleGAN synthesis network progressively convolves and upscales a low-resolution feature map into a high-resolution one. To condition the generator on a camera, we concatenate these feature maps with an appropriately downsampled Pl\"ucker embedding of the sampled camera parameters. By doing so, the GAN learns to associate camera rays with appearance.
}
  \label{fig:method}
\end{figure*}

\textit{Geometry-free} methods traditionally learn view consistency priors by training image synthesizers on large \textit{multi-view} datasets rather than using a 3D representation. LFNs~\cite{Sitzmann2021LFNs} and 3DiM~\cite{Watson2022NovelVS} are two successful methods that have inspired our approach. LFNs represents each scene as a light field parametrized by a multilayer perceptron (MLP) which maps a ray to a color. For generation, the MLP is also conditioned on a randomly-sampled latent code through meta-learning~\cite{ha2017hypernetworks}. However, compared to a convolutional neural network (CNN), a MLP cannot effectively utilize the inductive bias of spatial smoothness in natural images. The generated results are oftentimes blurry compared to CNN-based image synthesizers. 3DiM uses a more powerful image generator---a diffusion model~\cite{SohlDickstein2015DeepUL,ho2020denoising,song2021scorebased}, and achieve state-of-the-art single-view novel view synthesis results on the ShapeNet dataset~\cite{Shapenet2015}. At their core is a pose-conditional image-to-image diffusion model trained using ground-truth multi-view images as supervision. Being a conditional image synthesizer, 3DiM cannot perform unconditional multi-view image generation. While both works have made strong methodological contributions, neither has been proven to generate results at a resolution higher than $128\times 128$, to learn without multi-view datasets, nor to learn over photo-realistic images. In comparison, we show that our method (ray conditioning and a CNN synthesizer) can be trained with only \textit{single-view} posed data at $1024\times1024$ resolution, and can perform even better than \textit{geometry-based} methods on practical viewpoint-editing applications.

\section{Method}
Like prior work on 3D-aware GANs~\cite{Chan2021}, we focus on unstructured single-view image collections of objects from the same category, e.g., human faces, with labeled camera poses. Namely, we focus on a set of $(\rmI, \rmK, \rmE)$ triples, where $\rmI$ is an image, $\rmK$ are its corresponding intrinsics, and $\rmE = [\rmR\mid \rvt]$ are its corresponding camera extrinsics (camera-to-world transformations). For the sake of downstream applications such as portrait reposing, we seek to train a GAN that allows us to explicitly control the viewpoints of the synthesized images---without explicitly modeling the geometry. We accomplish this by adding our proposed ray-conditioning mechanism to an off-the-shelf image generator: StyleGAN2~\cite{Karras2019AnalyzingAI}. Our method requires minimal modifications to the image generator's architecture, while 
achieving higher photo-realism than methods that use a 3D representation~\cite{Chan2021,zhao-gmpi2022,orel2022stylesdf,gu2021stylenerf}.

\subsection{Photo Collections as Unstructured Light Fields}
 Images are often regarded as a sample of a scene's 5D plenoptic function, $L: (\mathbf{p}, \mathbf{d})\mapsto \mathbf{c}$, which describes the light intensity $\mathbf{c}$ in an arbitrary direction $\mathbf{d}\in \mathbb{S}^2$, and at arbitrary location $\mathbf{p} \in \mathbb{R}^3$~\cite{Adelson1991ThePF}. As described in the classic light field works~\cite{Levoy1996LightFields,Gortler1996Lumigraph}, if we choose to sample images outside a convex hull surrounding the object, the 5D plenoptic function becomes 4D. In this case, the 4D plenoptic function is also called the 4D light field. A pixel $(u, v)$ of an image $\rmI$ can be interpreted as a sample of the light through a ray direction:
 \begin{align}
 &\mathbf{d}_{u, v}=\mathbf{R}\mathbf{K}^{-1}\begin{bmatrix}u,v,1\end{bmatrix}^T ,\\
&\mathbf{d}_{u, v}=\mathbf{d}_{u, v} \big/ \lVert \mathbf{d}_{u, v} \rVert_2,\\
&u,v\in [0, W)\times [0, H), 
 \end{align}
 where $W,H$ are $\rmI$'s width and height.  A perspective image is then a measurement of the 4D light field at the location of the camera origin, i.e., $\mathbf{p}=\mathbf{o}$, from a bundle of ray directions $\mathbf{d}_{u, v}$ falling inside the camera's field-of-view. %
      
 Typically, the light field of a single scene is measured using a dense grid of synchronized cameras~\cite{Wilburn2005CameraArrays}. Novel views can then be synthesized by interpolating these densely captured images. 
 However, for a photo collection with only one image per scene, such as one of faces, the light field measurements are highly unstructured. Due to the varying identities, expressions etc, each image can be thought of as being a single-shot sampling of its scene's light field.

 The task of generative modeling is thus to model a distribution of light field observations over an entire photo collection to picture what missing view points may have looked like.

\subsection{Ray Conditioning for Image Synthesis}
Given a posed image collection $\big\{(\mathbf{I}, \mathbf{K}, \mathbf{E})\big\}$ 
that measures light fields in a highly unstructured way, we aim to learn the distribution of a light field $L$ defined on ray bundles $\mathbf{r} \in \mathbb{R}^{6\times H \times W}$. $\mathbf{r}$ is a 2D feature map which assigns each pixel $(u, v)$ in $\mathbf{I}$ to the camera ray through that pixel, $\mathbf{r}_{u, v}$, as illustrated in Figure~\ref{fig:method}. We only consider ray bundles that follow the same distribution as all the input ray bundles; hence $L$ is not well-defined for out-of-distribution ray bundles, as shown in Figure~\ref{fig:trans-limitation}.

We add ray conditioning to a GAN $G$ to model the distribution of a light field $L$ implicitly: the generator maps a Gaussian-distributed noise code $\mathbf{z}$ and ray bundle $\mathbf{r}$ to an image sample: $G(\rvz, \rvr) = \rmI$. By fixing the noise code $\mathbf{z}$, and using different ray bundles $\mathbf{r}$, we can sample images at different viewpoints from the same learnt light field. Concretely, we use the StyleGAN2 backbone, as does EG3D~\cite{Chan2021}. As shown in Figure~\ref{fig:method}, we add ray conditioning to each level of the progressively-growing synthesis network. At each level, we downsample the spatial ray embedding using bilinear interpolation, then concatenate the ray embedding and feature map together along the channel dimension.

\newcolumntype{b}{F}
\newcolumntype{s}{>{\hsize=.25\hsize}F}
\begin{figure*}[t]

  \newcommand{\imgw}{0.97}

  \setlength{\tabcolsep}{2.0pt}
  \begin{center}
  \scalebox{0.82}{
  \begin{tabular}{ll}
    \rotatebox{90}{\hspace{14.0pt} EG3D \cite{Chan2021}} & \includegraphics[width=\imgw\linewidth]{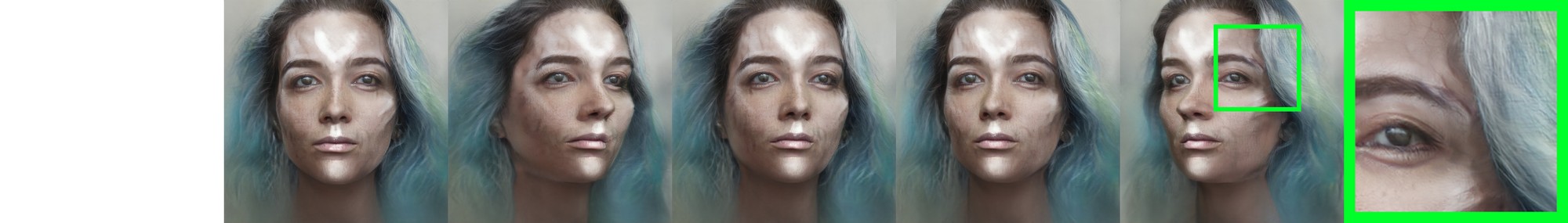} \\
    \rotatebox{90}{\hspace{12.0pt} GMPI \cite{zhao-gmpi2022}} & \includegraphics[width=\imgw\linewidth]{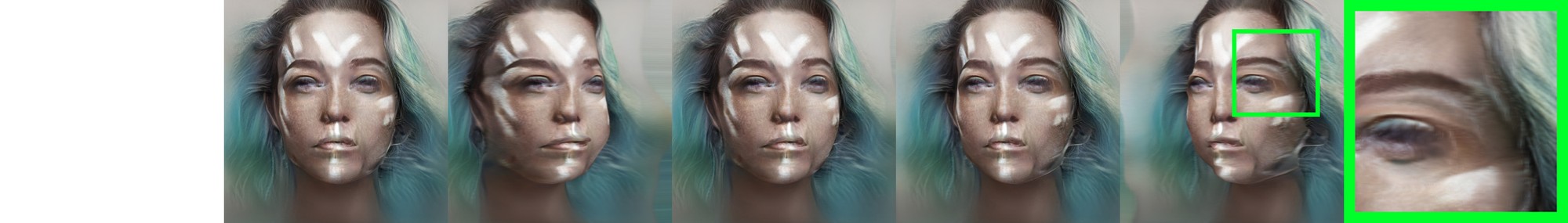} \\
    \rotatebox{90}{\hspace{10.0pt} Ray Cond.} & \includegraphics[width=\imgw\linewidth]{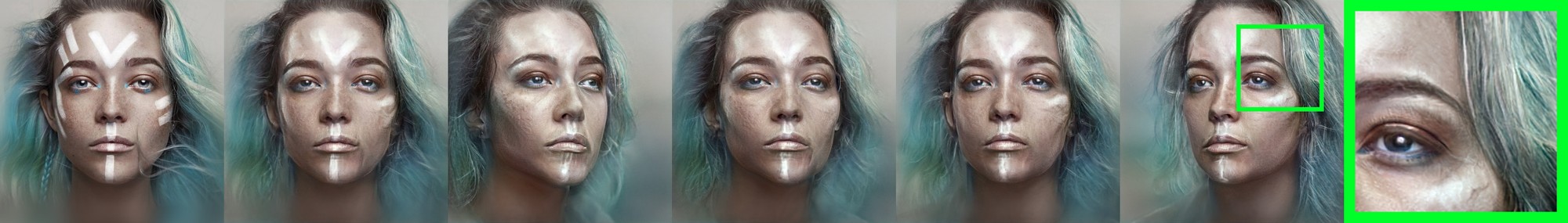} \\
    \rotatebox{90}{\hspace{14.0pt} EG3D \cite{Chan2021}} & \includegraphics[width=\imgw\linewidth]{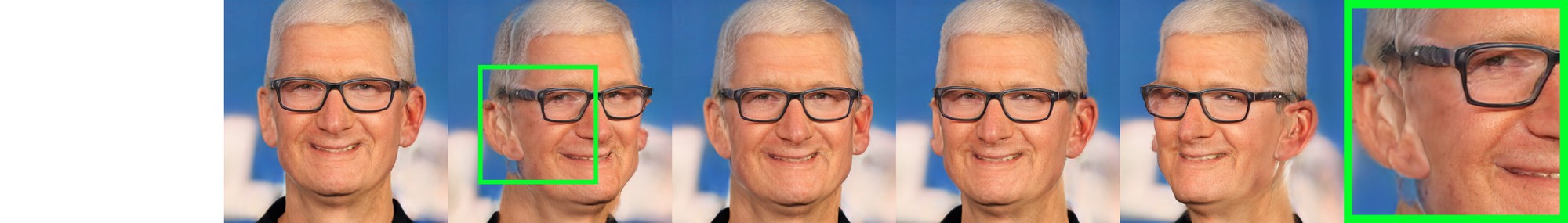} \\
    \rotatebox{90}{\hspace{12.0pt} GMPI \cite{zhao-gmpi2022}} & \includegraphics[width=\imgw\linewidth]{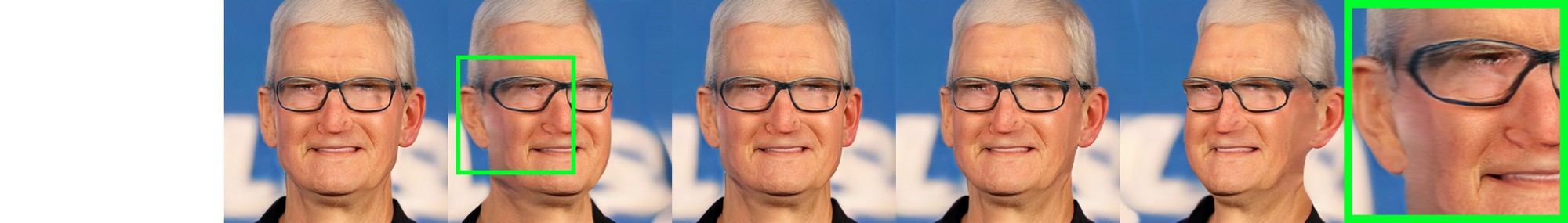} \\
    \rotatebox{90}{\hspace{10.0pt} Ray Cond.} & \includegraphics[width=\imgw\linewidth]{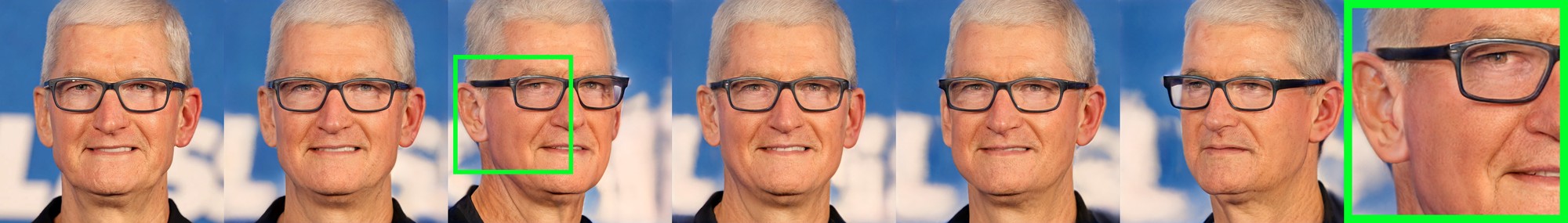}
  \end{tabular}
  }
  \end{center}
  \vspace{-10.0pt}
  \scalebox{0.8}{
  \hspace{96.0pt} Input \hspace{34.0pt} Inversion \hspace{132.0pt} Rotation
  }

  \caption{\textbf{Viewpoint Editing via GAN Inversion.} We invert an image into each GAN using PTI~\cite{roich2021pivotal}. Ray conditioning is able to best preserve the details of the input images, as shown in the rich detail of the skin, eyes, and hair. Because radiance fields are biased towards low frequency results, the EG3D~\cite{Chan2021} inversions lack the detail that ray conditioning can offer. In the second example, there are also geometry artifacts near the ears and border of the face. GMPI~\cite{zhao-gmpi2022} struggles to invert input images, causing  distortion in the novel views.}
  \label{fig:inversion}
\end{figure*}

The ray embedding needs to be carefully chosen: the standard 5D ray parametrization $\rvr_{u, v} = (\mathbf{o},\mathbf{d}_{u, v})$ is redundant for a 4D light field as it fails to consider the assumption of zero decay in empty space: $L(\mathbf{o}, \mathbf{d})=L(\mathbf{o}+t\mathbf{d}, \mathbf{d})$.  Inspired by LFNs~\cite{Sitzmann2021LFNs}, we remove this redundancy through the Pl\"ucker parametrization $\rvr_{u, v} = (\mathbf{o}\times \mathbf{d}_{u, v},\mathbf{d}_{u, v})$, where $\times$ is the cross product. With this parametrization, we have:
 \begin{equation}
    \left(\rvo  + t\rvd \right) \times \rvd  = \rvo \times \rvd + t\rvd \times \rvd = \rvo \times \rvd.
\end{equation}
 
\begin{figure*}[h!]
  \centering
    \setlength{\tabcolsep}{2.0pt}

    \newcommand{\figqrw}{0.46}
    \begin{tabular}{llll}
    \rotatebox{90}{\hspace{-34.0pt} Pose Cond. \cite{Chan2021}} & \includegraphics[width=\figqrw\linewidth]{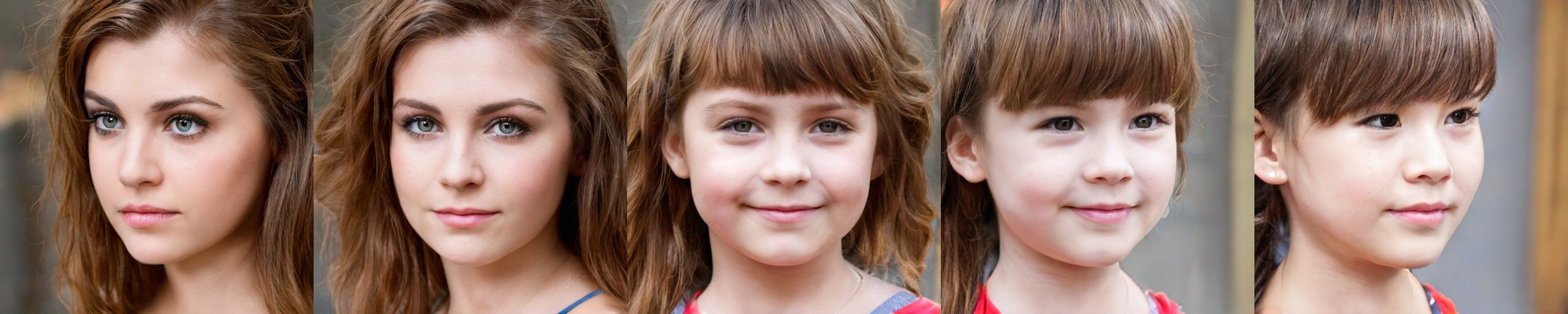} & \rotatebox{90}{\hspace{-24.0pt} EG3D \cite{Chan2021}} & \includegraphics[width=\figqrw\linewidth]{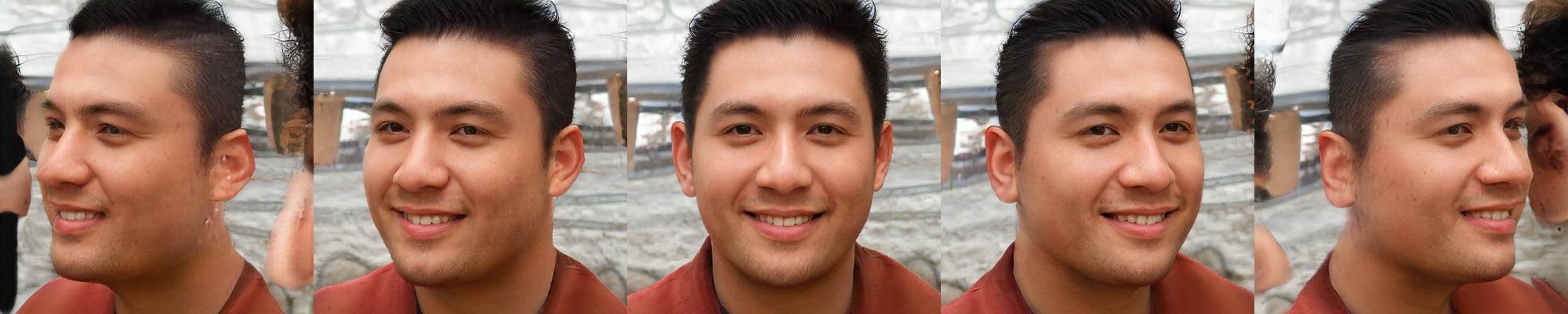} \\
    & \includegraphics[width=\figqrw\linewidth]{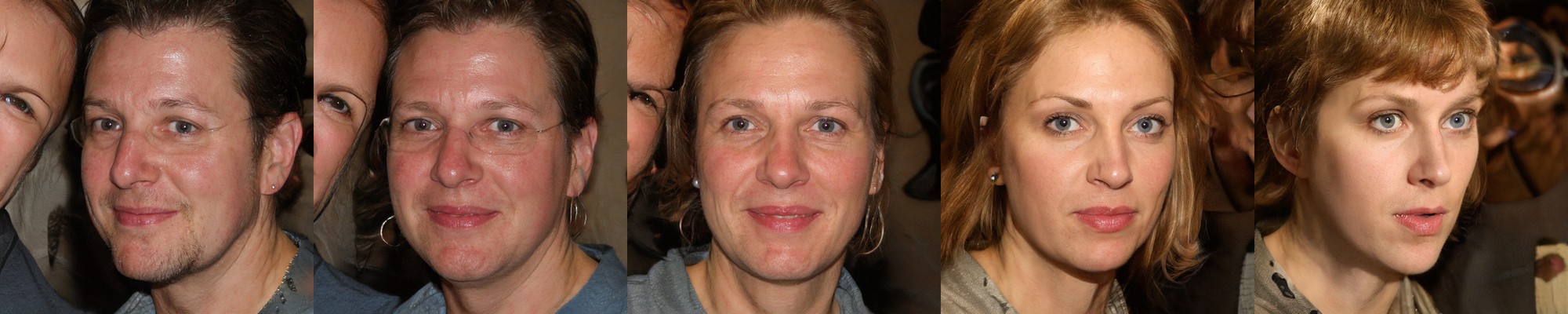} & & \includegraphics[width=\figqrw\linewidth]{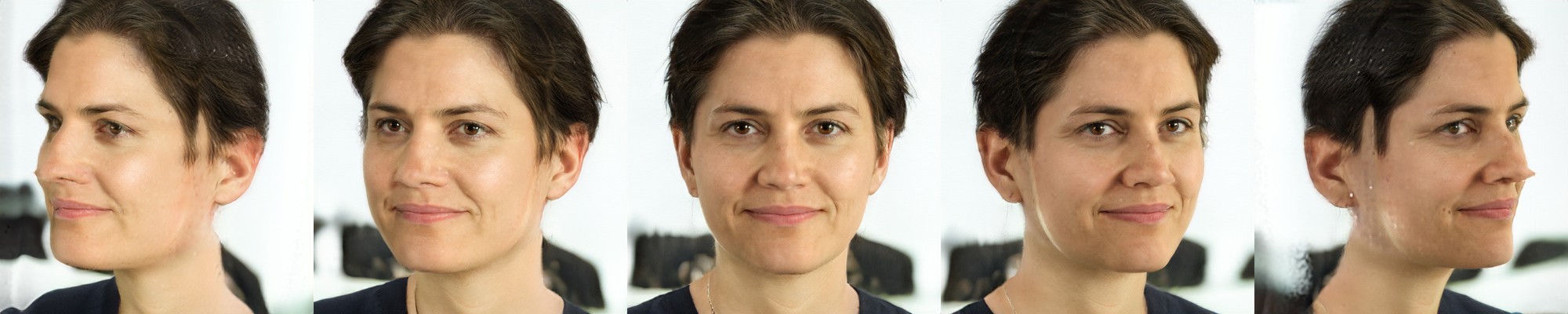} \\
    \rotatebox{90}{\hspace{-26.0pt} GMPI~\cite{zhao-gmpi2022}} & \includegraphics[width=\figqrw\linewidth]{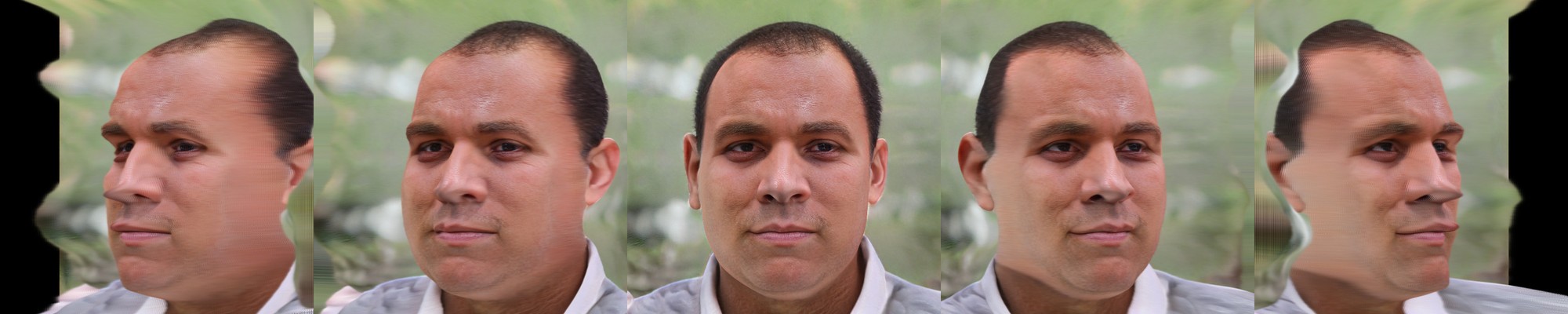} & \rotatebox{90}{\hspace{-26.0pt} Ray Cond.} & \includegraphics[width=\figqrw\linewidth]{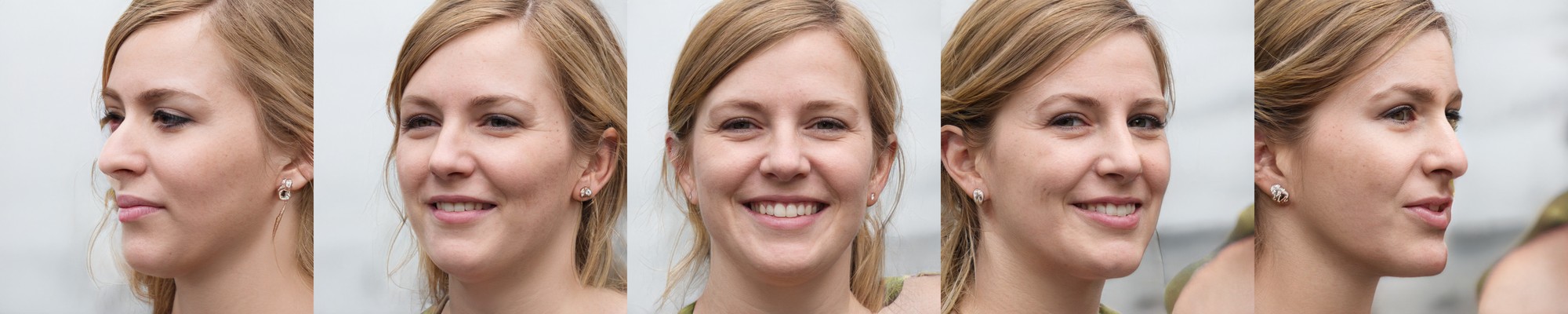} \\
    & \includegraphics[width=\figqrw\linewidth]{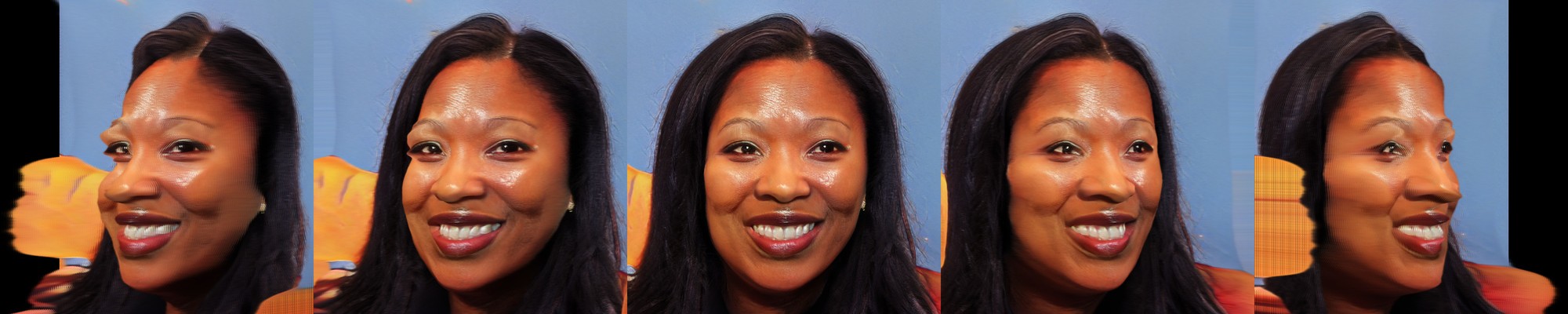} & & \includegraphics[width=\figqrw\linewidth]{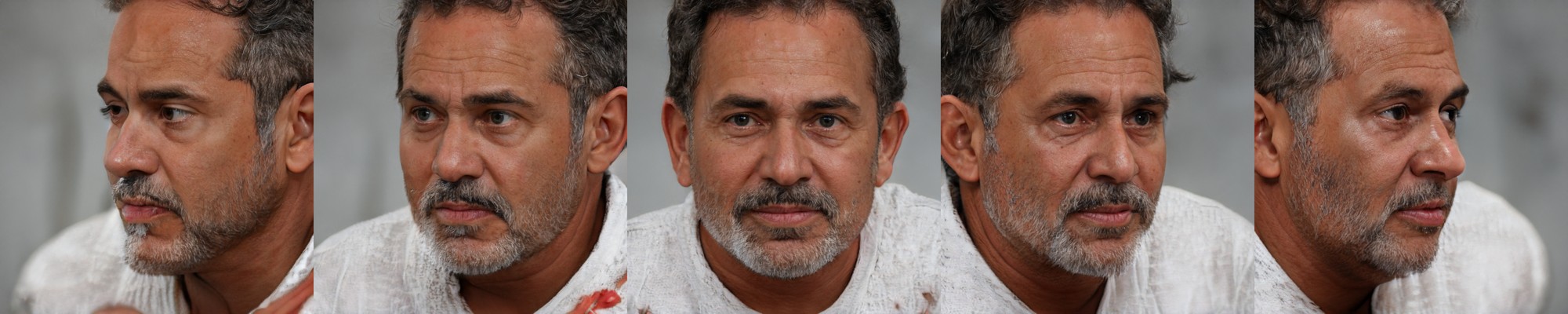} \\
    & \hspace{10.0pt} $-45^\circ$ \hspace{20.0pt} $-23^\circ$ \hspace{24.0pt} $0^\circ$ \hspace{22.0pt} $+23^\circ$ \hspace{18.0pt} $+45^\circ$ & & \hspace{10.0pt} $-45^\circ$ \hspace{20.0pt} $-23^\circ$ \hspace{24.0pt} $0^\circ$ \hspace{22.0pt} $+23^\circ$ \hspace{18.0pt} $+45^\circ$ 
    \end{tabular}

  \caption{\textbf{Unconditional Multi-view Image Generation.} As reflected in the photos, our method is able to maintain identity consistency and photorealism across angles. Pose conditioning fails to be view consistent. By foregoing a 3D model, we do not have the geometric artifacts that GMPI~\cite{zhao-gmpi2022} and EG3D~\cite{Chan2021} may have. Although there are some view-dependent changes between images, we achieve the highest image quality at steep angles. All results are generated with $\psi_{\text{trunc}}=0.7$.}
  \label{fig:yaw}
\end{figure*}

We require minimal modifications to the backbone StyleGAN2 architecture: each convolution kernel just needs to accept extra ray embedding inputs. Hence the induced computational overhead is almost negligible. Moreover, we can start from a pretrained StyleGAN2 model, and finetune it to make it amenable to explicit viewpoint control. Unlike prior works that prioritize photo-consistency over photo-realism, our method maintains the high image generation fidelity of StyleGAN2, as shown in Figure~\ref{fig:yaw}.

\subsection{Viewpoint Editing for Real Posed Images}
\label{sec:viewpointediting}
For generated images, new viewpoints can be achieved by changing the ray bundles $\rvr$ in our generated light field $G(\rvz, \rvr)$. Like prior work in StyleGAN-based real image editing, we can invert a real posed image $(\mathbf{I}, \rmK, \rmE)$ into a latent space of StyleGAN first, and then modify the ray bundles to edit the viewpoint. 

As our method closely resembles the backbone StyleGAN2, we can directly use off-the-shelf GAN inversion methods~\cite{roich2021pivotal} for high-quality inversions and viewpoint edits. To invert the image's camera parameters, we use Deep 3D Face Reconstruction~\cite{deng2019accurate}. This is in stark contrast to geometry-based methods like EG3D that require more dedicated inversion methods, as shown by concurrent work~\cite{xie2022HFGI3D} and Figure~\ref{fig:inversion}. 

 We also differ from prior latent-space viewpoint editing work in terms of offering intuitive explicit viewpoint control and increased viewpoint change range. We provide results in the supplementary material. In addition, those latent-space editing directions require paired training data~\cite{shen2020interfacegan, styleflow}. One such method, InterfaceGAN \cite{shen2020interfacegan}, relies on an external binary classifier to determine whether a generated face is facing left or right. InterfaceGAN also does not allow for explicit control of pose, and relies on manual tuning to achieve the desired pose. 

\section{Experiments}

\label{sec:experiments}
We validate our approach by testing its ability to generate multi-view images on two single-view posed datasets: Flickr-Faces-HQ (FFHQ)~\cite{Karras2018ASG} and AFHQv2 Cat Faces~\cite{choi2020starganv2}. We show that our geometry-free approach outperforms geometry-based baselines in terms of photo-realism when generating multi-view images. We also demonstrate our model's strength for downstream viewpoint editing  of real images. Finally,  we compare our method with a prior geometry-free method, LFNs~\cite{Sitzmann2021LFNs}, on a multi-view posed dataset: SRN Cars~\cite{sitzmann2019srns,Shapenet2015}, and show significant improvement in multi-view generation quality.

\subsection{Metrics}
We adopt the same metrics used in EG3D~\cite{Chan2021} for quantitative evaluation. 

\noindent\textbf{Image Quality.} To compare image generation quality, we report the FID score~\cite{Heusel2017GANsTB} and KID score$\times 100$~\cite{Binkowski2018DemystifyingMG} between 50k generated images and the entire training set. %

\noindent\textbf{Identity Consistency.} As a proxy for view consistency, we use ArcFace~\cite{deng2018arcface,serengil2020lightface}, a facial recognition model, to compute identity consistency between two random views of one individual, and average it over 1024 samples. %

\noindent\textbf{Pose Accuracy.} We sample one camera pose for 1024 individuals and estimate the camera pose of the synthesized image with Deep 3D Face Reconstruction~\cite{deng2019accurate}. We report the mean squared error between camera angles. %

\noindent\textbf{Generation Speed.} We benchmark the time for generating an image from a latent vector and report the generation speed in frames per second (FPS). %
Generation speed is critical for GAN inversion, as inversion methods may require hundreds of queries to the GAN to invert a real image. %

\subsection{Results and Discussion}

We report quantitative results comparing ray conditioning to competitive 3D-aware baselines in Table~\ref{tab:3d_aware}, and show samples of generated images in Figure~\ref{fig:yaw}. All baseline results except for FPS are quoted from StyleGAN2-ADA~\cite{Karras2020ada}, GMPI~\cite{zhao-gmpi2022} and EG3D~\cite{Chan2021}. We also report standard deviations for the ID metric and pose metric, which were previously not reported. Notably, we are able to achieve strong results with only a 2D GAN backbone. Because our image quality is not limited by the resolution of a geometric model, ray conditioning is able to achieve the highest FID and KID scores on faces, which are most similar with the scores of StyleGAN2. We are also able to achieve competitive identity accuracies and pose accuracies, demonstrating our model's ability to maintain view consistency and model camera pose. 

\begin{table*}[t!]
\centering
\scalebox{0.9}{
\begin{tabular}{lrrrrrrr}
\toprule
    & \multicolumn{5}{c}{FFHQ} &  \multicolumn{2}{c}{AFHQv2 Cats}  \\
& FID$\downarrow$ & KID$\downarrow$ & ID $\uparrow$ & Pose$\downarrow$ & FPS$\uparrow$ & FID$\downarrow$  & KID $\downarrow$ \\
          \midrule
    \textbf{Image synth.} \\
    StyleGAN2~\cite{Karras2020ada} $512^2$ & - & - &- & - & 69 & $3.55^\dagger$ & 0.066$^\dagger$ \\
    StyleGAN2~\cite{Karras2019AnalyzingAI} $1024^2$ & 2.70 & 0.048 &- & - & 62 & -& -\\
    \midrule
    \textbf{Geometry-based MV synth.}\\
     StyleSDF~\cite{orel2022stylesdf} $256^2$ & 11.5 &0.370 & - & -&- &$12.8^*$&$0.447^*$\\
     StyleNeRF~\cite{gu2021stylenerf} $1024^2$ & 8.10 & 0.240 & - & - &- & $14.0^*$ &$0.350^*$\\
    EG3D~\cite{Chan2021} $512^2$ &4.70 & 0.132 & \ $\mathbf{0.77}$\small{$\pm 0.15$} & $\mathbf{0.005}$\small{$\pm 0.005$} &33 & $\mathbf{2.77}^\dagger$ & $\mathbf{0.041}^\dagger$\\
     GMPI~\cite{zhao-gmpi2022} $512^2$ & 8.29 & 0.454 &  $\mathbf{0.74}$\small{$\pm 0.16$} &$\mathbf{0.006}$\small{$\pm 0.009$} &13& 7.79 &0.474\\
      GMPI~\cite{zhao-gmpi2022} $1024^2$ & 7.50 & 0.407 &  $\mathbf{0.75}$\small{$\pm 0.16$} &$\mathbf{0.007}$\small{$\pm 0.010$} &$6$& - &-\\
      \midrule
     \textbf{Geometry-free MV synth.}\\
     Ray Conditioning $512^2$ & $3.50$ &$0.076$& $\mathbf{0.75}$\small{$\pm 0.15$} & $\mathbf{0.006}$\small{$\pm 0.007$}&$\mathbf{48}$& 3.44 & 0.103\\
     Ray Conditioning $1024^2$ & $\mathbf{3.28}$ &$\mathbf{0.066}$& $\mathbf{0.76}$\small{$\pm 0.14$} & $\mathbf{0.006}$\small{$\pm 0.007$} &$\mathbf{38}$& - & -\\
     \bottomrule\\
\end{tabular}
}

\caption{\textbf{Multi-view Image Generation Metrics.} Ray conditioning enables multi-view (MV) image synthesis by conditioning a 2D GAN on a ray embedding of a camera. It achieves high degrees of photorealism, identity consistency, and pose accuracy. We compare each multi-view GAN method to a StyleGAN2 baseline, showing the loss of fildelity due to geometric inductive biases. All metrics except for FPS are quoted from StyleGAN2-ADA~\cite{Karras2020ada}, EG3D~\cite{Chan2021} and GMPI~\cite{zhao-gmpi2022}. We also compute and report standard deviations for the ID and pose scores, which were previously not reported. We bold the best statistically significant results. *Trained on all of AFHQ instead of the cats subset. $^\dagger$Trained with adaptive discriminator augmentation~\cite{Karras2020ada}.
}
\label{tab:3d_aware}
\end{table*}

These metrics corroborate what we see in Figure~\ref{fig:yaw}. Compared to EG3D and GMPI, ray conditioning maintains the highest visual quality across yaw changes. GMPI appears to have lower quality images than EG3D and ray conditioning. As reported in the GMPI ablation study, we believe that this is due to GMPI's synthetic shading, which is necessary to learn accurate depth information. At yaws of $\pm 23^\circ$, all methods do a good job at representing rotation. However, beyond that, EG3D and GMPI appear to show artifacts. For EG3D, there is noise near the ears in both examples. For the second individual, there is an unnatural border near the ears as well. This is a known failure mode of EG3D and other radiance fields called billboarding, described in Section 1.3 of their appendix. For GMPI, the multiplane images cause noticeable quality issues in the rendered images. Images appear to be blurry. While ray conditioning sacrifices some view consistency, such as changes in smile, the results remain realistic even at difficult yaws. These results provide an explanation for why ray conditioning is able to achieve the best FID and KID scores. Forfeiting a 3D representation allows for high quality image synthesis across angles. The fact that all of the most competitive methods are built upon the StyleGAN architecture underscores its superior ability to disentangle pose and appearance. Ray conditioning is a natural extension of StyleGAN for generating multi-view images. 

\begin{figure}[h]
  \centering

  \setlength{\tabcolsep}{2.0pt}
  \begin{tabular}{ll} 
  \rotatebox{90}{LFNs~\cite{Sitzmann2021LFNs}} & \includegraphics[width=0.94\linewidth]{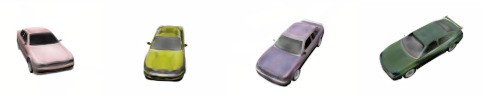} \\ 
  \rotatebox{90}{Ray Cond.} & \includegraphics[width=0.95\linewidth]{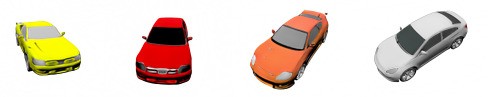} 
  \end{tabular}

  \caption{\textbf{Unconditional Generation of Cars}. 
  We observe that ray conditioning, a GAN-based method, has sharper images and more diverse samples than the baseline LFNs~\cite{Sitzmann2021LFNs}. In particular, ray conditioning achieves an FID of $3.39$ whereas LFNs achieves an FID of $41.8$. (FID is computed from $50$k random samples.)
  }
  \label{fig:cars-comparison}
\end{figure}

In Figure~\ref{fig:cars-comparison}, we compare ray conditioning to Light Field Networks~\cite{Sitzmann2021LFNs} (LFNs), a model designed for geometry-free view synthesis. LFNs use an autodecoder to condition the color of a ray on a normally distributed latent vector. Because LFNs are trained with a L2 reconstruction loss instead of an adversarial loss, output images tend to be blurry. Relative to the training dataset, LFNs achieves an FID score of $41.8$ while ray conditioning achieves an FID score of $3.39$. The samples from ray conditioning are much sharper, and show more diversity. We provide videos in the supplementary material. As shown in Figure~\ref{fig:face-comparison}, on FFHQ, LFNs struggles to reconstruct the input data. It is also not able to synthesize novel views when trained on FFHQ, a single-view dataset. Ray conditioning demonstrates that light field conditioning concept introduced in LFNs is capable of synthesizing compelling results on only single-view data.

\subsection{Ablation Study}
\label{sec:ablation}
 We compare ray conditioning to a simpler alternative: pose conditioning. Similar to StyleGAN2-ADA~\cite{Karras2020ada} and EG3D~\cite{Chan2021},
 we first flatten the camera extrinsics $\rmE\in \mathbb{R}^{4\times 4}$ and intrinsics $\rmK\in \mathbb{R}^{3\times 3}$ into a conditioning vector $\rvc \in \mathbb{R}^{25}$. We then input both a randomly sampled $\rvz\sim \mathcal{N}(0, 1)^{512}$ and the conditioning vector $\rvc$ into the mapping network for predicting a $\rvw$ code, as shown in Figure~\ref{fig:method}.
 Pose conditioning also provides explicit viewpoint control; however, it encodes a viewpoint as a 1D vector, rather than 2D feature map as we do for ray conditioning. We show that the lack of spatial inductive bias in pose conditioning causes the identity of generated people to vary wildly with small rotations in Figure~\ref{fig:yaw}. 
 At a resolution of $512\times512$, pose conditioning achieves an average ID similarity score of $0.68\pm 0.20$, while ray conditioning achieves $0.75\pm 0.15$. Ray conditioning produces much more view-consistent results than pose conditioning.
\begin{figure}[h]
  \centering

  \scalebox{0.86}{
  \setlength{\tabcolsep}{2.0pt}
  \begin{tabular}{ll} 
  \rotatebox{90}{\hspace{6.0pt}LFNs~\cite{Sitzmann2021LFNs}} & \includegraphics[width=0.95\linewidth]{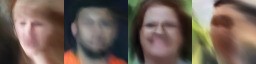} \\ 
  \rotatebox{90}{\hspace{6.0pt} Ray Cond.} & \includegraphics[width=0.95\linewidth]{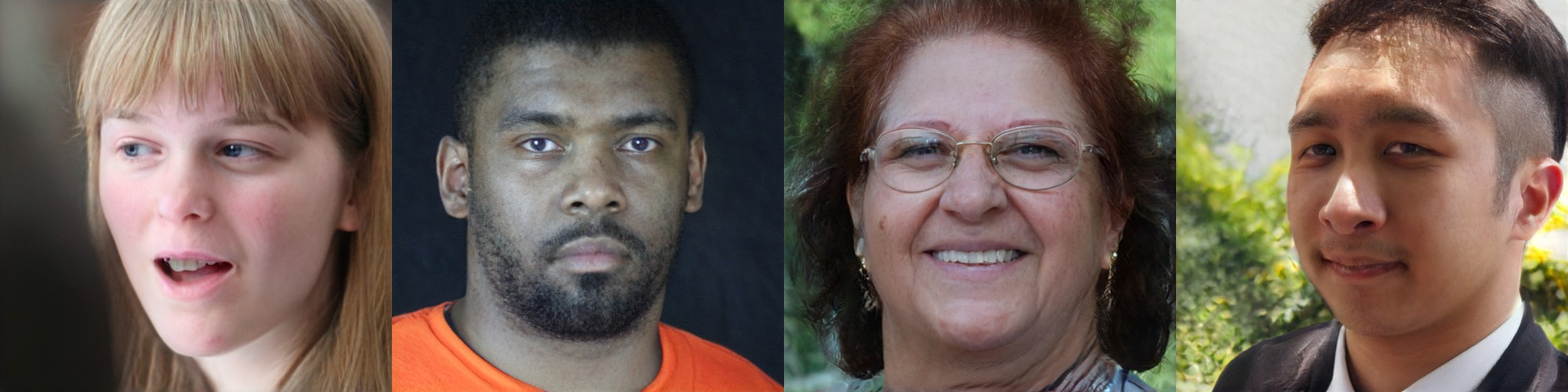} \\
  \rotatebox{90}{\hspace{16.0pt} Input} & \includegraphics[width=0.95\linewidth]{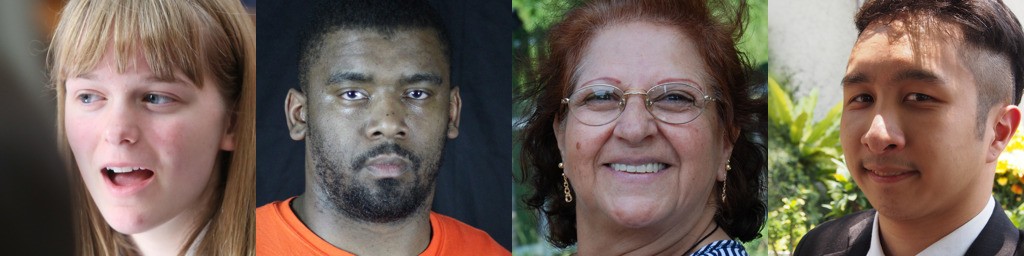}
  \end{tabular}
  }

  \caption{\textbf{Face Reconstruction Comparison}. 
  LFNs~\cite{Sitzmann2021LFNs} struggle to reconstruct the training data of FFHQ, making it unsuitable for generating light fields from a \textit{single-view} dataset of natural images. Moreover, it is not able to synthesize novel views. Ray conditioning is able to successfully reconstruct the training images with GAN inversion.   
  }
  \label{fig:face-comparison}
\end{figure}

\subsection{Viewpoint Editing for Real Posed Images}
We compare our ray conditioning against prior work for the application of editing viewpoints of real posed images, as described in Section \ref{sec:viewpointediting}. Since our method and baselines are all variants of StyleGAN, we invert images using Pivotal Tuning Inversion (PTI)~\cite{roich2021pivotal}, and synthesize images of the same individual from different viewpoints. We show in Figure~\ref{fig:inversion} that our ray conditioning method achieves the same explicit viewpoint control as the geometry-based methods while preserving a much higher degree of photo-realism. In the first example, the detail is incredibly noticeable in the eyes and hair, which closely resemble the input image. Eyes are especially important for human perception of identity and familiarity, but are  often difficult to invert for geometry-based methods due to their specularity.  

Moreover, geometry-based methods such as EG3D and GMPI can introduce geometric artifacts in synthesized images. 
When fitting a geometry-based representation to a single image, there is often ambiguity on whether to modify the geometry or texture. Incorrect geometry can create seemingly correct images. This is only realized after a shift of viewpoint. For radiance fields, this has been coined as shape radiance ambiguity~\cite{kaizhang2020}, and still a challenging problem for many 3D representations. In the bottom individual of Figure~\ref{fig:inversion}, we see that although EG3D is able to reproduce the input image, the disoccluded parts around the ears exhibit strong geometry artifacts when viewed at a different yaw angle. GMPI is more severely hurt by the ambiguity between geometry and appearance when fitting to the input image, which leads to distortion in the novel views. This is most noticeable in the bottom individual's glasses, which appear to be glued to the face. Additionally, since geometry-based methods tend to smooth textures in favor of photo-consistency, they lack the level of details that our ray conditioning can offer. Rich details in hair, skin, and eyes are inherently view-dependent, and are best captured by relaxing constraints on photo-consistency.

Furthermore, we quantitatively evaluate ray conditioning and EG3D on GAN inversion and viewpoint editing with the CelebA-HQ~\cite{CelebAMask-HQ} dataset. Neither method was trained on this dataset, allowing for a measure of cross-dataset generalization. To evaluate the similarity between input images and inversions, we calculate PSNR, SSIM, LPIPS~\cite{zhang2018perceptual}, and ID scores. To evaluate the image quality of synthesized novel views, we compare the FID and KID$\times100$ against the original images. Both metrics were computed from one novel viewpoint for $100$ images at a resolution of $512\times512$. The results in Table~\ref{tab:inversion} show that ray conditioning can achieve higher image quality and detail preservation in both input inversions and after viewpoint editing.

\begin{table}[h]
\centering
\scalebox{0.75}{
\begin{tabular}{lrrrrrr}
\toprule
&  \multicolumn{4}{c}{Inversion} &  \multicolumn{2}{c}{Novel Views}\\
& PSNR$\uparrow$ & SSIM$\uparrow$ & LPIPS $\downarrow$ & ID$\uparrow$ & FID$ \downarrow$ & KID $\downarrow$\\
          \midrule
     EG3D~\cite{Chan2021} & 26.56 & $\mathbf{0.78}$ & 0.12 & $0.75$ & 65.0 & 0.0180\\
     Ray Cond. &  $\mathbf{27.49}$ & $\mathbf{0.78}$ & $\mathbf{0.10}$ & $\mathbf{0.85}$ & $\mathbf{58.5}$ & $\mathbf{0.0036}$\\
     \bottomrule\\
\end{tabular}
}

\caption{\textbf{GAN Inversion Metrics.} We measure reconstruction quality between input images and inversions on four metrics. We then compare the FID and KID$\times100$ between the original images and images with random viewpoint changes. Ray conditioning can achieve higher image quality and detail  in both input inversions, and after viewpoint change.}
\label{tab:inversion}
\end{table}

\begin{figure}[h]
  \centering

  \setlength{\tabcolsep}{1.0pt}
  \scalebox{1}{
  \begin{tabular}{ll}
    \rotatebox{90}{\hspace{6.0pt} Trans.} & \includegraphics[width=0.92\linewidth]{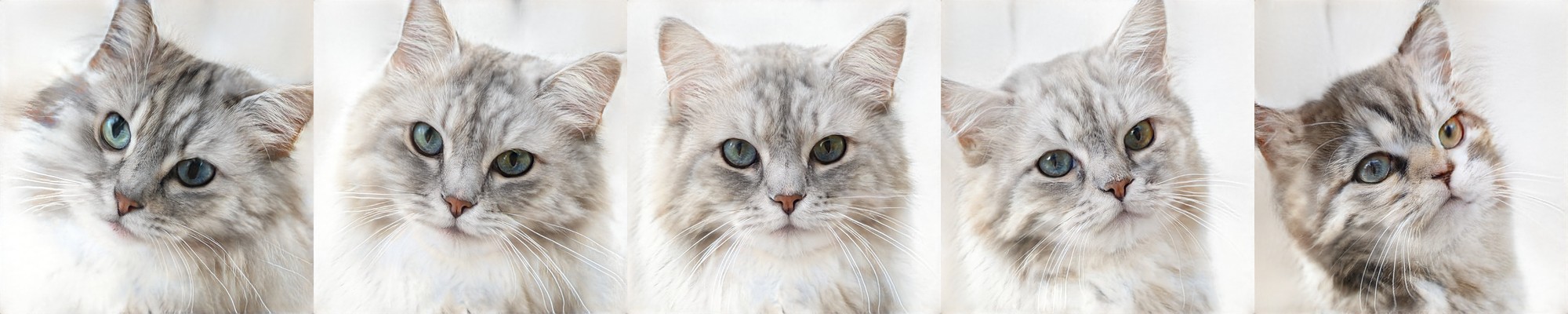} \\
    \rotatebox{90}{\hspace{10.0pt} Rot.} & \includegraphics[width=0.92\linewidth]{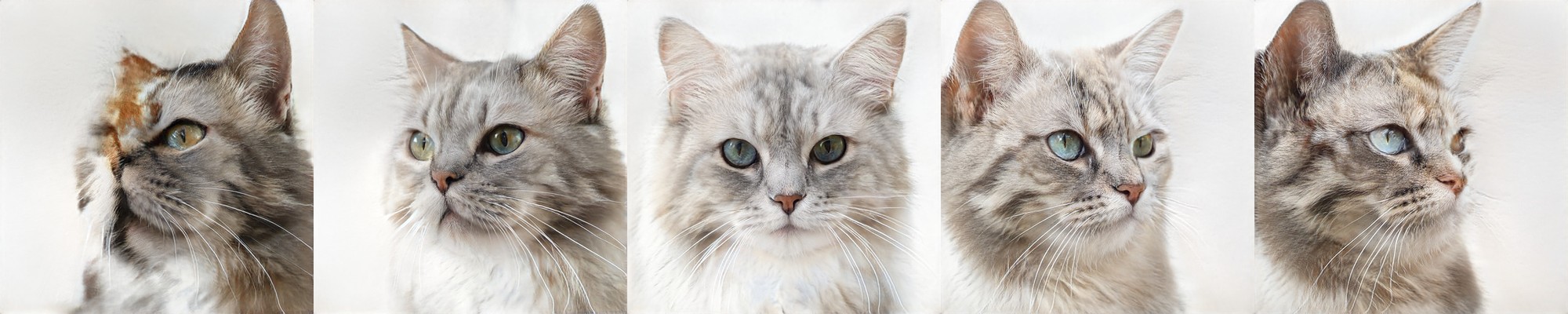}
  \end{tabular}
  }

  \caption{\textbf{Limitations on Camera Control}. Ray conditioning does not generalize well to out-of-distribution camera poses. The first row shows that $\pm 1.0$ of $x$-axis camera translation leads to viewpoint distortion and identity shift. The second row shows that an out of distribution rotation of $\pm 75^\circ$ leads to identity shift.
  }
  \label{fig:trans-limitation}
\end{figure}

\section{Conclusion}
We propose ray conditioning, a method for multi-view image generation with explicit viewpoint control. %

 Our key insight is that we do not need to generate consistent 3D geometry to control the viewpoint of generated images. Instead, 4D ray conditioning lets us generate different viewpoints individually, placing fewer constraints on the generator, which leaves it freer to optimize for photo-realism. However, through our experiments, we find that this comes with a trade-off. While ray conditioning creates realistic static images, it may introduce aliasing in videos. It also does not generalize well to out-of-distribution camera poses, as shown in Figure~\ref{fig:trans-limitation}.
 
The difference between EG3D and ray conditioning echoes that of 3D geometry-based representations and light fields. If a subject is perfectly photo-consistent, then all views of the subject can be perfectly encoded in a 3D set of RGBA points. However, view-dependent effects such as specularities violate this assumption, as does high-frequency geometry when 3D resolution is finite~\cite{Chai2000Plenoptic}. The 4D light field accommodates such features by representing rays individually, which lets light reflected from a shared 3D point vary with angle.

By conditioning a 2D GAN on a light field prior, as opposed to using a 3D representation, we achieve the best photo-realism among all existing multi-view image synthesizers, with competitive identity consistency across viewpoints. We believe that, our method pushes forward the boundary of geometry-free generative models, and hope our conclusions can inspire a variety of work in new scene representations.

{\small
\bibliographystyle{ieee_fullname}
\bibliography{egbib}
}

\clearpage
\newpage
{\Large\textbf{Appendix}}

\appendix

In this document, we present implementation details and additional results that are supplemental to the main paper. In Appendix~\ref{app:design}, we outline key design details for the ray conditioning method. In Appendix~\ref{app:additional_results}, we include more results of viewpoint editing. We also compare to concurrent work in 3D-aware image inversion, and show that these methods still have challenges with respect to realism. Finally, in Appendix~\ref{app:eval}, we include more details about the datasets used.  

\section{Design Details}
\label{app:design}
We outline some key design details in this section. 

\subsection{Pretraining and Weight Initialization}
\begin{figure}[h]
\centering

\includegraphics[width=\linewidth]{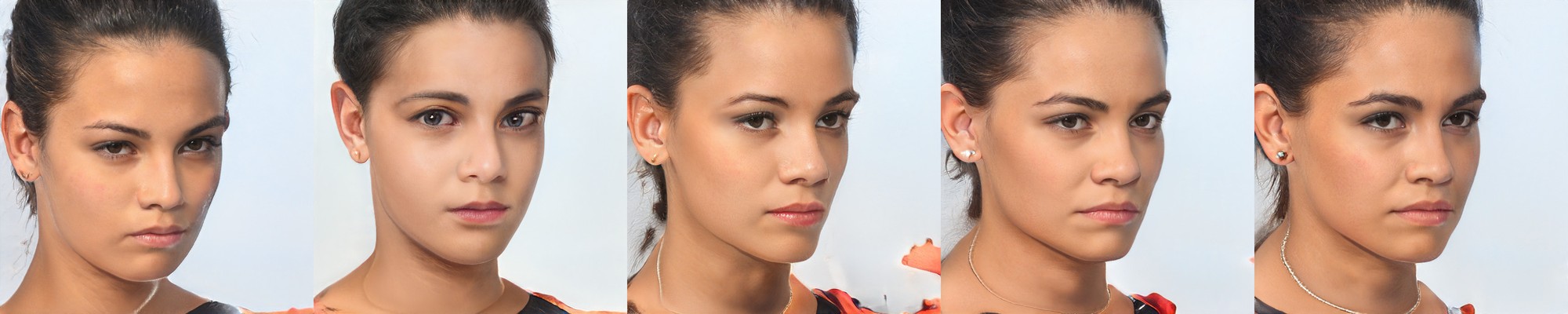}
\includegraphics[width=\linewidth]{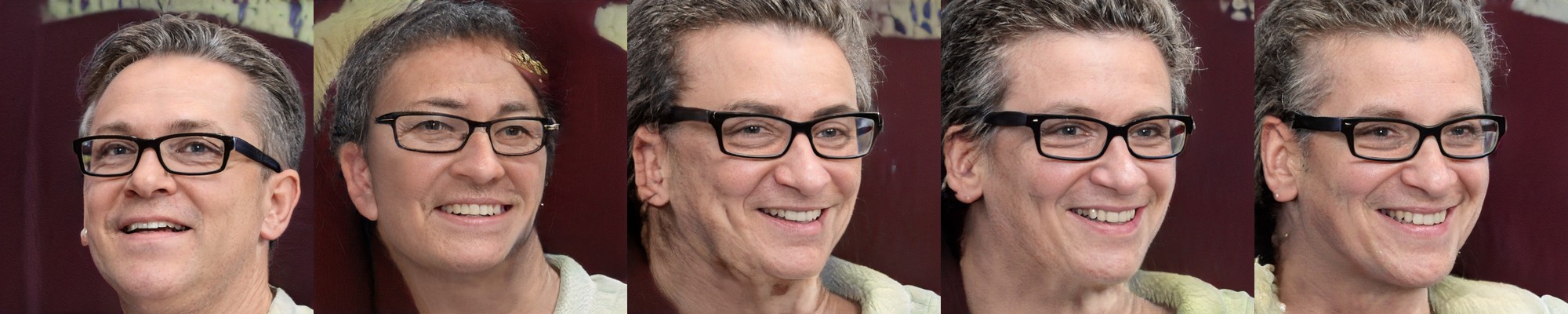}
\begin{tabularx}{\linewidth}{FFFFF}
Init & 80kimgs & 160kimgs& 240kimgs & 320kimgs
 \end{tabularx}

\caption{\textbf{Training Convergence Speed.} For faces, we initialize our ray conditioning model from a pretrained StyleGAN2 model. Through training, it learns to properly generate images from a target pose. Ray conditioning converges quickly. After 160kimgs (1.5 hours on 2 A6000s), ray conditioning is already able to properly generate an image at a target pose. }
\label{fig:convergence}
\end{figure}

\begin{figure*}
\centering
\includegraphics[width=\linewidth]{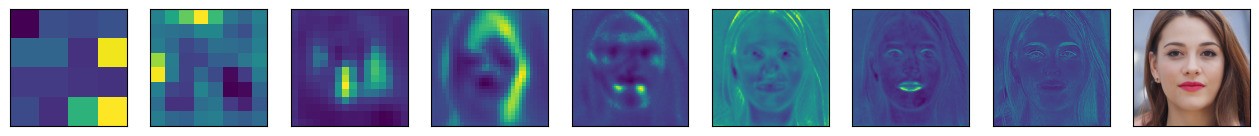}
\includegraphics[width=\linewidth]{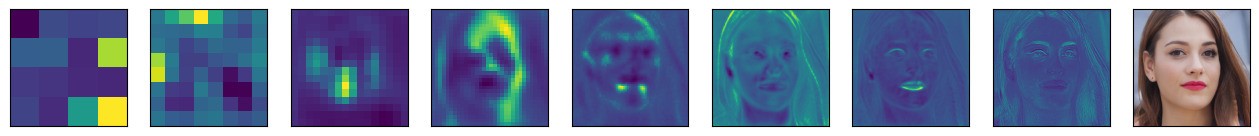}
\includegraphics[width=\linewidth]{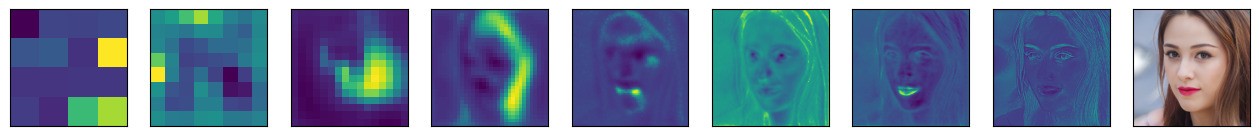}

\begin{tabularx}{\linewidth}{FFFFFFFFF}
$4\times4$ & $8\times8$ & $16\times16$ & $32\times32$ & $64\times64$ & $128 \times 128$& $256\times 256$ & $512\times 512$ & RGB
\end{tabularx}

\caption{\textbf{Ray Conditioning Affects Intermediate Features.} To learn viewpoint control, we condition each intermediate layer of StyleGAN on a ray embedding. We picture coarse to fine resolution feature maps of one latent seed from different poses. Consistent with prior work~\cite{Wu2020StyleSpaceAD,harkonen2020ganspace}, we find that subject pose is a coarse-level feature in the StyleGAN latent space. It begins to take form around resolutions $16\times 16$ to $64\times 64$. }
\label{fig:intermediate_full}
\end{figure*}

Recall that the ray embedding is a $6 \times H \times W$ feature map which is concatenated to each intermediate representation of StyleGAN. To accommodate these extra features, we add 6 channels to each convolutional layer of StyleGAN. To begin training, we initialize all prior weights to be those of a pretrained StyleGAN model. The extra 6 channels for each layer is then initialized normally, with the default StyleGAN2 gain parameters~\cite{Karras2019AnalyzingAI,He2015DelvingDI}. The discriminator also uses pretrained discriminator weights. The pose conditioninal discriminator module is initialized as an MLP in the same way done in EG3D and StyleGAN2-ADA~\cite{Chan2021,Karras2020ada}. In Figure~\ref{fig:intermediate_full}, we demonstrate the effect that ray conditioning has on the intermediate layers of StyleGAN. Consistent with prior work~\cite{Wu2020StyleSpaceAD,harkonen2020ganspace}, subject pose is generally a coarse-level feature which takes form around resolutions $16\times 16$ to $64\times 64$. Ray conditioning also converges very fast. In Figure~\ref{fig:convergence}, we show that ray conditioning takes about 160kimgs, or 1.5 hours on 2 Nvidia A6000s, to learn camera pose. 

\subsection{Effect on Latent Space}
Many prior work have found directions in StyleGAN's $\mathcal{W}$ latent space which correspond to subject pose~\cite{harkonen2020ganspace,shen2020interfacegan}. Surprisingly, ray condition nullifies these directions in the latent space. When we try to modify an image's pose using InterfaceGAN directions~\cite{shen2020interfacegan} after ray conditioning, the image stays the same. This implies that ray conditioning ``moves" the pose information previously embedded in the $\mathcal{W}$ latent space into the convolutional weights assigned to the ray embedding. 

\subsection{StyleGAN2 vs. StyleGAN3.} We have implemented ray conditioning successfully for both StyleGAN2~\cite{Karras2019AnalyzingAI} and StyleGAN3~\cite{Karras2021AliasFreeGA}. StyleGAN3 is better suited for video generation tasks because of its antialiasing abilities. However, for multi-view image generation, we have found no advantage for using StyleGAN3. Following EG3D~\cite{Chan2021} and GMPI~\cite{zhao-gmpi2022}, we choose to use StyleGAN2 because of its slightly higher image quality and faster model. 

\subsection{Training Details}
FFHQ and AFHQ models were trained starting from official StyleGAN checkpoints. They were trained on $2\times$ Nvidia A6000 GPUs for 1040kimgs - 1440kimgs. For SRN Cars, we train from scratch for 13,520kimgs. Hyperparameters are set to the same as those of StyleGAN2. 

\section{Additional Results}
\label{app:additional_results}
\subsection{Viewpoint Editing Examples}
We believe that ray conditioning is a natural choice for portrait editing over 3D-aware GANs. In Figure~\ref{fig:photo_edit_eg3d}, we recreate Figure 2 in the main paper with EG3D. We see that the resulting images from EG3D appear more cartoonish than human-like. There are also geometry distortions. In the top right individual (yellow background), we see geometry artifacts near his ear. In the bottom example (pink background), the challenges are also apparent when we try to blend faces into a preexisting image. EG3D fails to achieve our intended effect of photorealistic viewpoint editing. In the supplementary material, we also include a video demonstrating the viewpoint control we have over input samples. 

\subsection{3D-Aware GAN Inversion}
Many have recognized the issues with using a 2D GAN inversion method such as PTI~\cite{roich2021pivotal} with a 3D-aware GAN such as EG3D~\cite{Chan2021}. Although PTI can successfully invert an input image, it can cause geometry artifacts that are only realized after a change in viewpoint. Several work have created dedicated 3D GAN inversion methods for EG3D. However, we find that these methods can still cause aliasing in the inverted images, creating a loss of quality. In addition, they can struggle when inverting side-facing images. The occluded regions can have incorrect geometry. The eyes can also lose their specularity, which is important for human perception of identity. In Figure~\ref{fig:3d_inversion}, and Figure~\ref{fig:3d_inversion_2} we demonstrate these issues. We compare ray conditioning to two related work. 3D GAN Inversion with Pose Estimation~\cite{ko20233d} is a recent work designed for EG3D inversion. HFGI3D~\cite{xie2022HFGI3D} is a concurrent work also designed for EG3D inversion. Inversion with dedicated 3D-aware methods can also take much longer than with PTI. For instance, 3D GAN Inversion with Pose Estimation~\cite{ko20233d} takes on average 3 minutes per image. HFGI3D~\cite{xie2022HFGI3D} can take 8 minutes. Meanwhile, PTI with ray conditioning only takes 1 minute per image.

\subsection{Latent Space Pose Editing}
InterfaceGAN and similar work~\cite{shen2020interfacegan,harkonen2020ganspace,styleflow} allow for viewpoint change by finding directions in the StyleGAN latent space which correspond to pose. However, these methods can only do binary changes such as left facing or right facing, instead of explicit viewpoint control. Because they do not operate per-pixel like how ray conditioning does, they also lack the spatial inductive bias which makes ray conditioning effective. We show an example of the differences between latent space editing and ray conditioning in Figure~\ref{fig:interfacegan}.

\subsection{Latent Space Samples}
In Figure~\ref{fig:latent_samples}, we show uncurated latent space samples from StyleGAN2 with ray conditioning. Even without a 3D representation, ray conditioning is still able change the viewpoint of generated samples. We picture latent seeds 0-31. To show our image quality, we also present larger images in Figure~\ref{fig:sample-results}.
\subsection{Additional Results on Cars}
\begin{figure}[h]
  \centering
  \includegraphics[width=\linewidth]{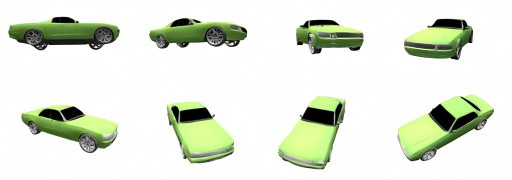}
  \includegraphics[width=\linewidth]{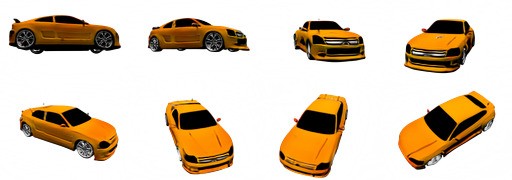}

  \caption{\textbf{$\mathbf{360^\circ}$ View Consistency from Ray Conditioning.} When trained on a multi-view dataset, ray conditioning is able to generate view consistent results with $360^\circ$ of rotation. Please see the accompanying videos for continuous results.}
  \label{fig:cars-view-consistency}
\end{figure}

In Figure~\ref{fig:cars-view-consistency}, we demonstrate that our model can enable $360^\circ$ viewpoint editing when trained on a dataset with $360^\circ$ of views. The car stays consistent as we rotate the camera. We also include videos of smooth trajectories in the supplementary material. We use StyleGAN3~\cite{Karras2021AliasFreeGA} because of its antialiasing properties. 

\subsection{More Experiments on Light Field Networks}
\begin{figure}[h]
\centering
 \rotatebox{90}{
 \begin{tabularx}{156pt}{FFF}
NVs & Recon. & Input
 \end{tabularx}
 }
\includegraphics[width=0.9\linewidth]{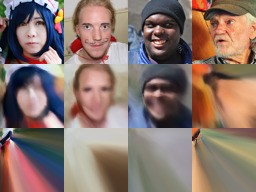}

\caption{\textbf{Light Field Networks on FFHQ.} Light Field Networks (LFNs)~\cite{Sitzmann2021LFNs} have two key challenges. First, they struggle to reconstruct high frequency details from input images. Second, when trained on a dataset with only one image per face, LFNs~\cite{Sitzmann2021LFNs} struggle to construct novel views (NVs). When combined with a more powerful generative model such as a GAN, ray conditioning helps to address both of these problems. }
\label{fig:lfns_faces}
\end{figure}

\begin{figure*}
\centering
\includegraphics[width=0.49\linewidth]{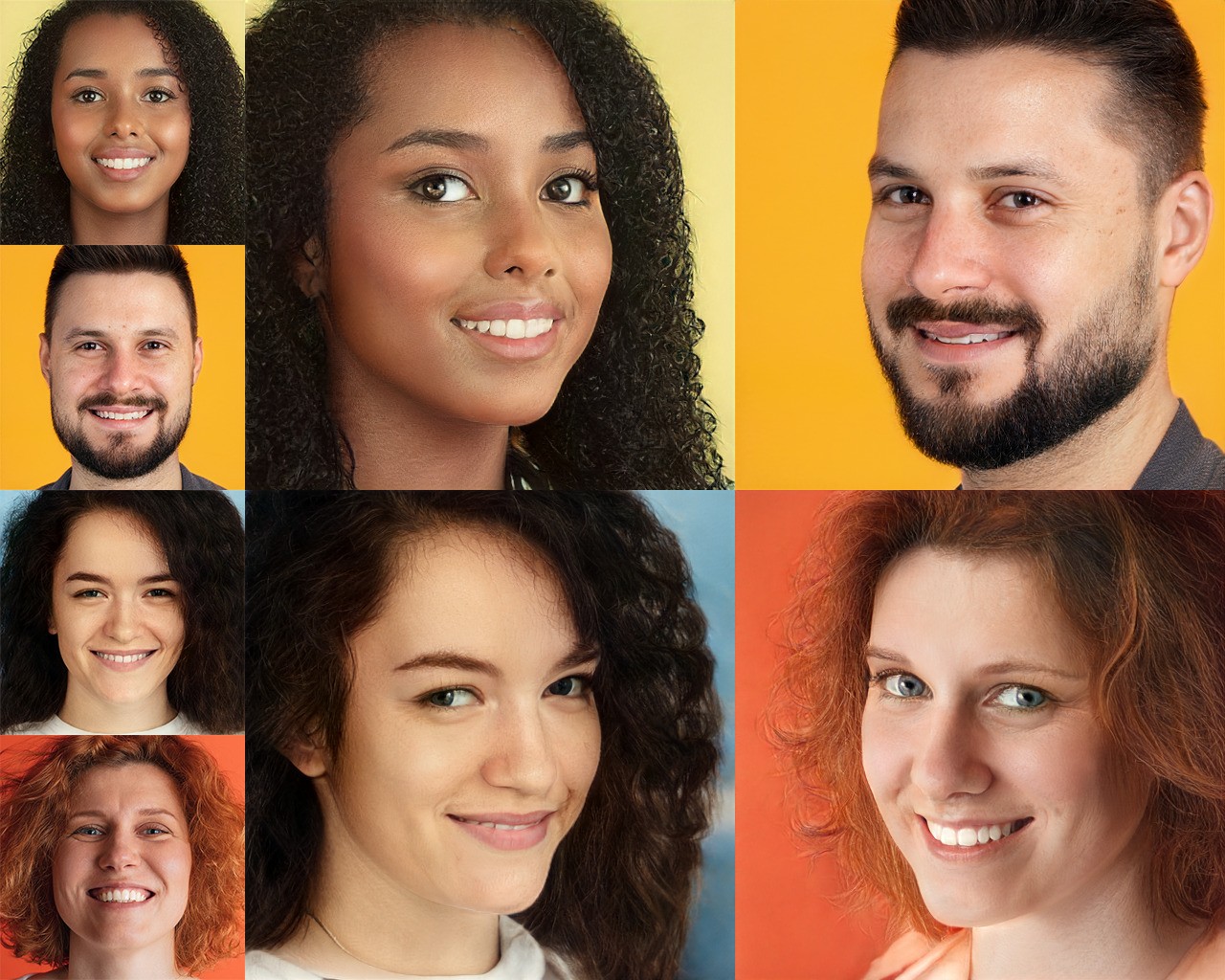}
\includegraphics[width=0.49\linewidth]{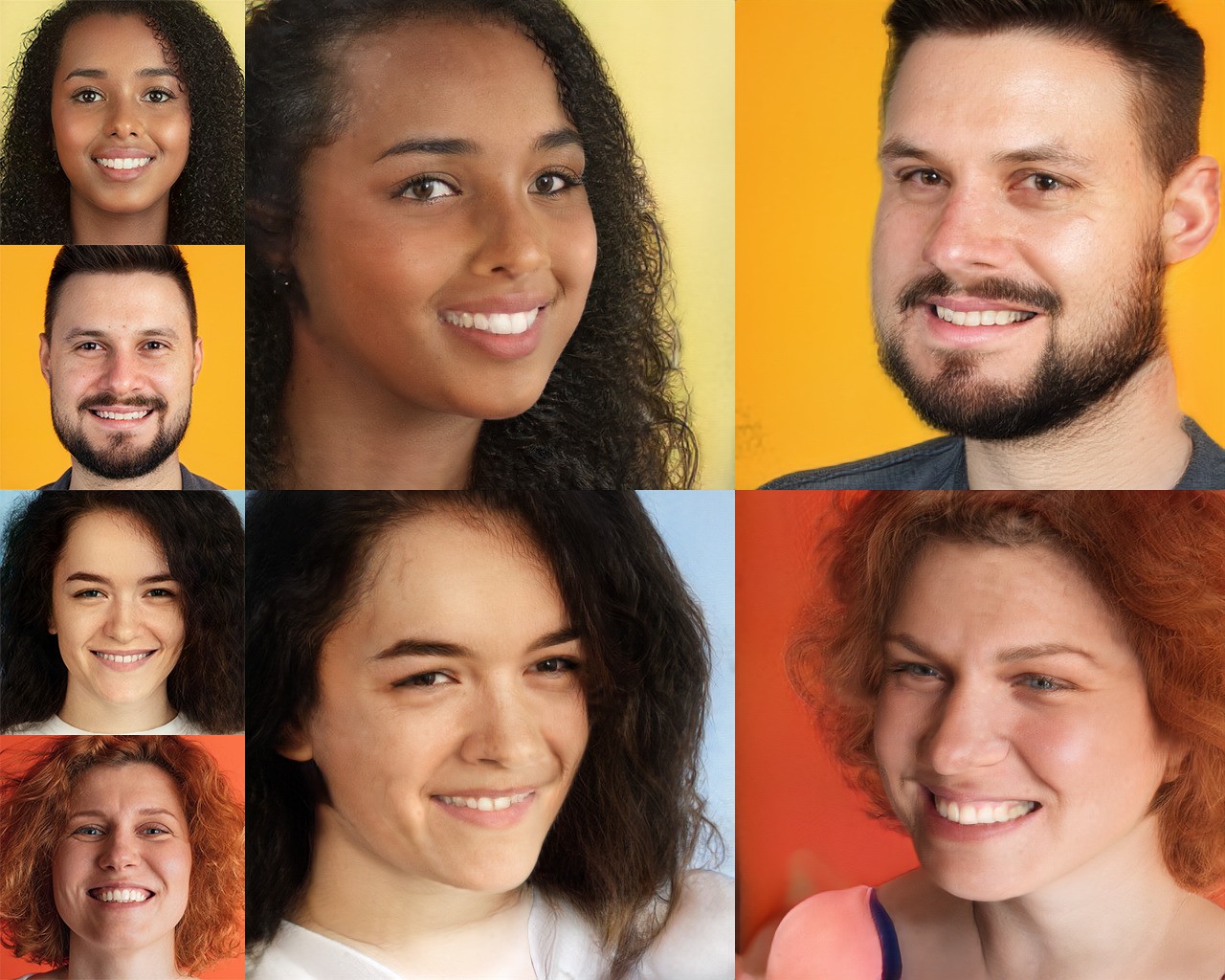}
\includegraphics[width=0.49\linewidth]{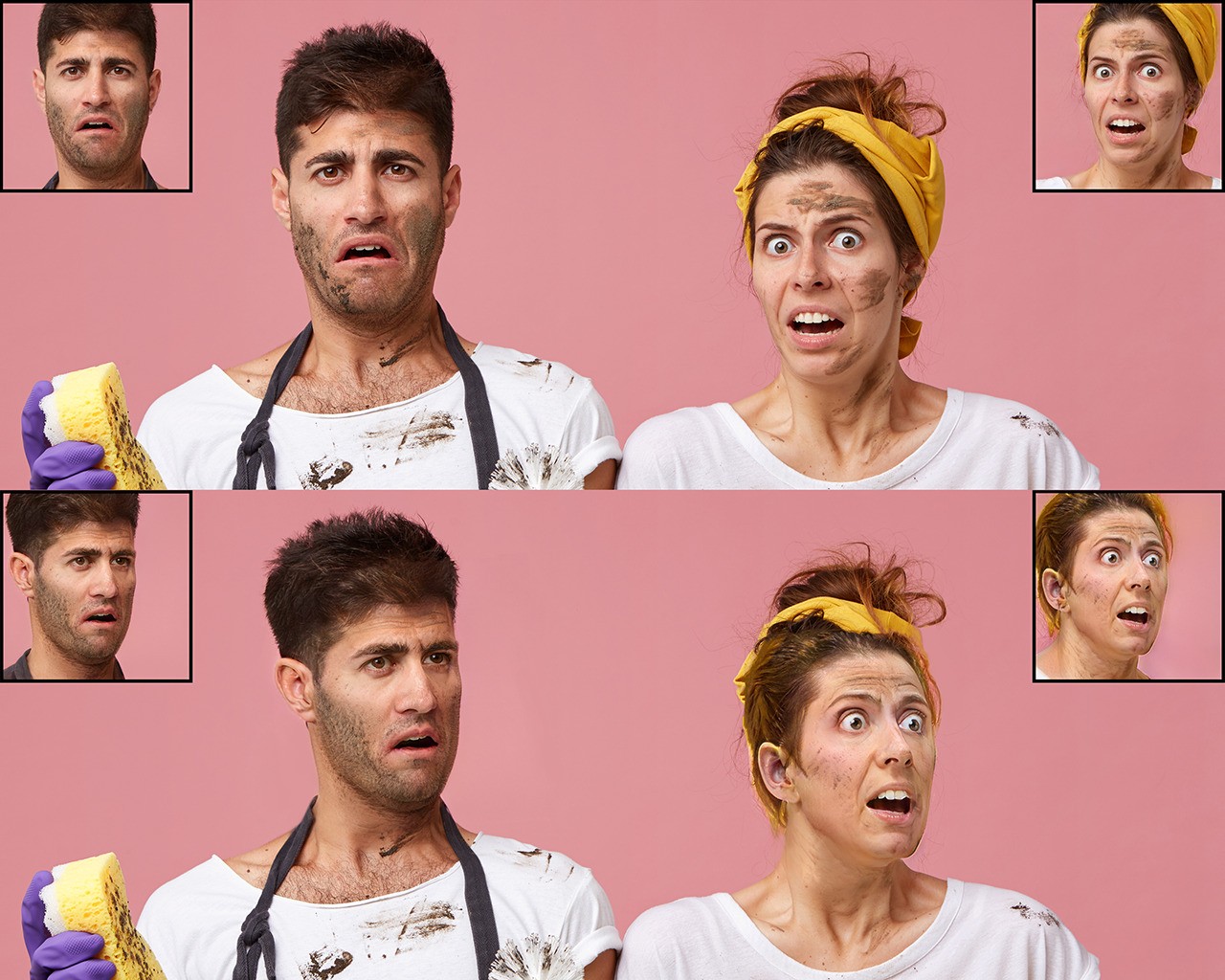}
\includegraphics[width=0.49\linewidth]{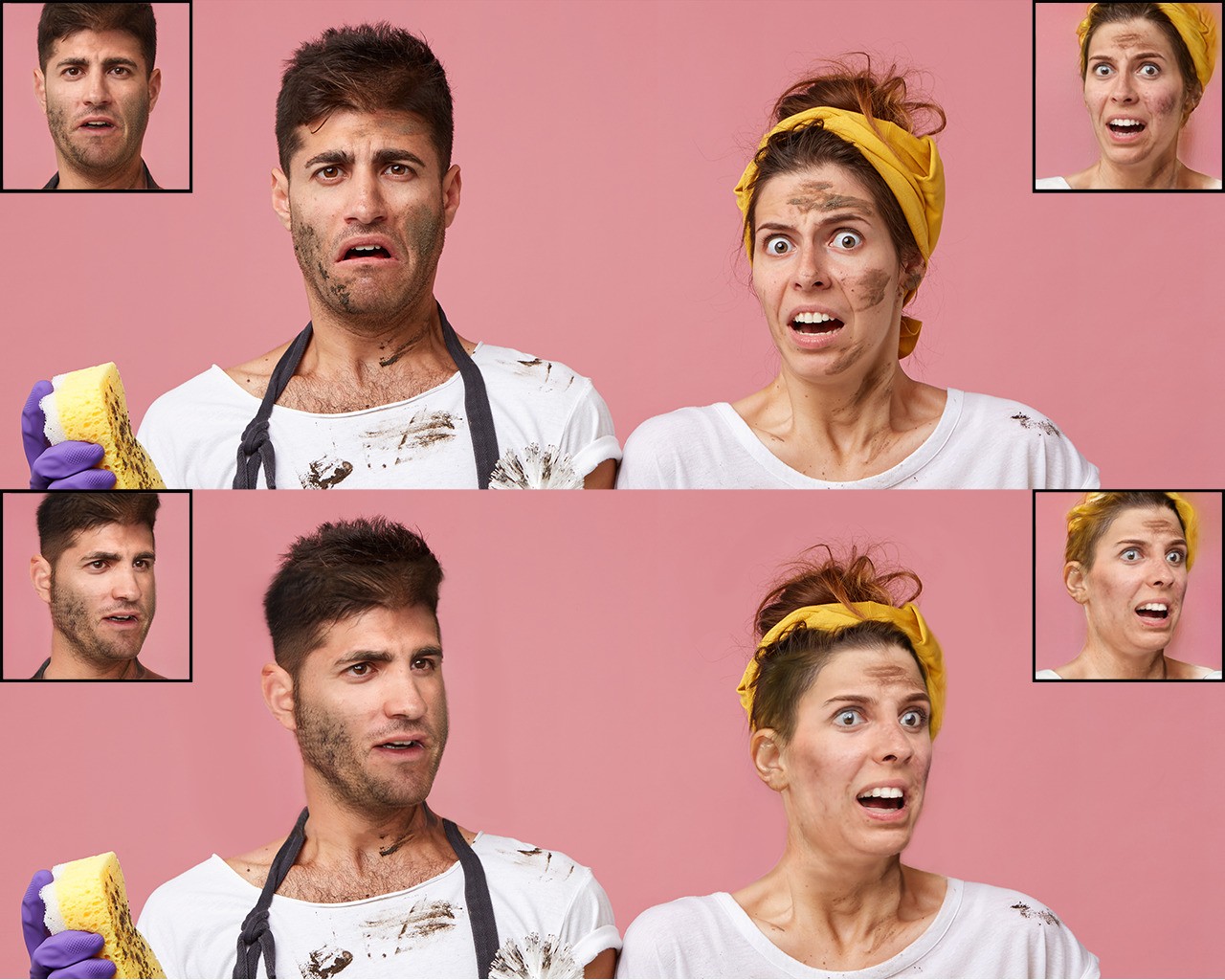}

\begin{tabularx}{\linewidth}{FF}
Ray Conditioning + PTI~\cite{roich2021pivotal} & EG3D~\cite{Chan2021} + PTI~\cite{roich2021pivotal}
\end{tabularx}

\caption{\textbf{Photo Editing Comparison.} We believe that ray conditioning is a natural choice for portrait editing over 3D-aware GANs. To illustrate, we compare the ray conditioning results from Figure 2 in the main paper to EG3D~\cite{Chan2021}. We see that EG3D cannot create a change in viewpoint without sacrificing some realism. There are noticeable geometry issues, and the edited individuals seem more cartoonish than human-like.}
\label{fig:photo_edit_eg3d}
\end{figure*}

\begin{figure*}
\centering
\newcolumntype{b}{F}
\newcolumntype{s}{>{\hsize=.3\hsize}F}
\includegraphics[width=0.495\linewidth]{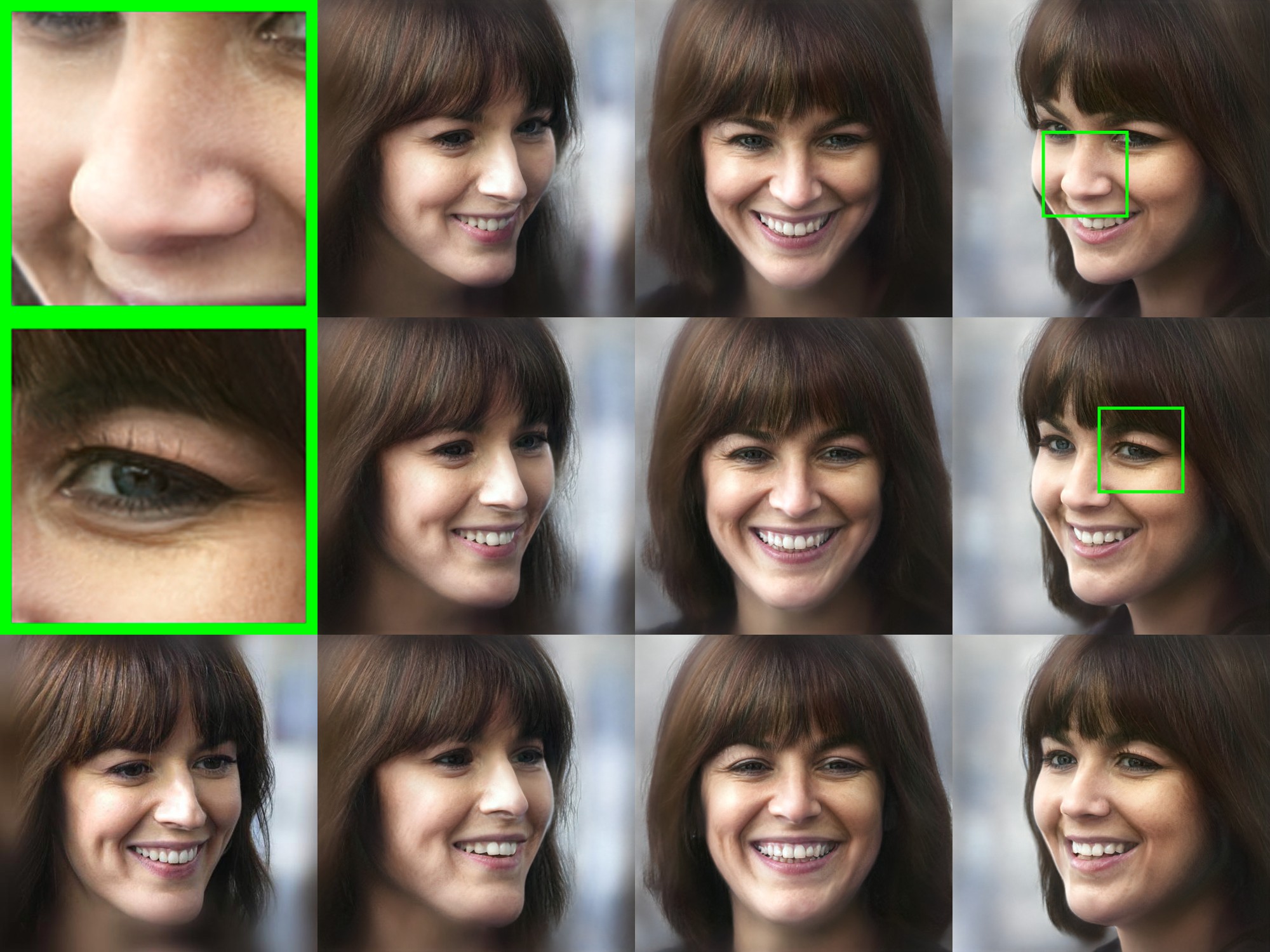}
\includegraphics[width=0.495\linewidth]{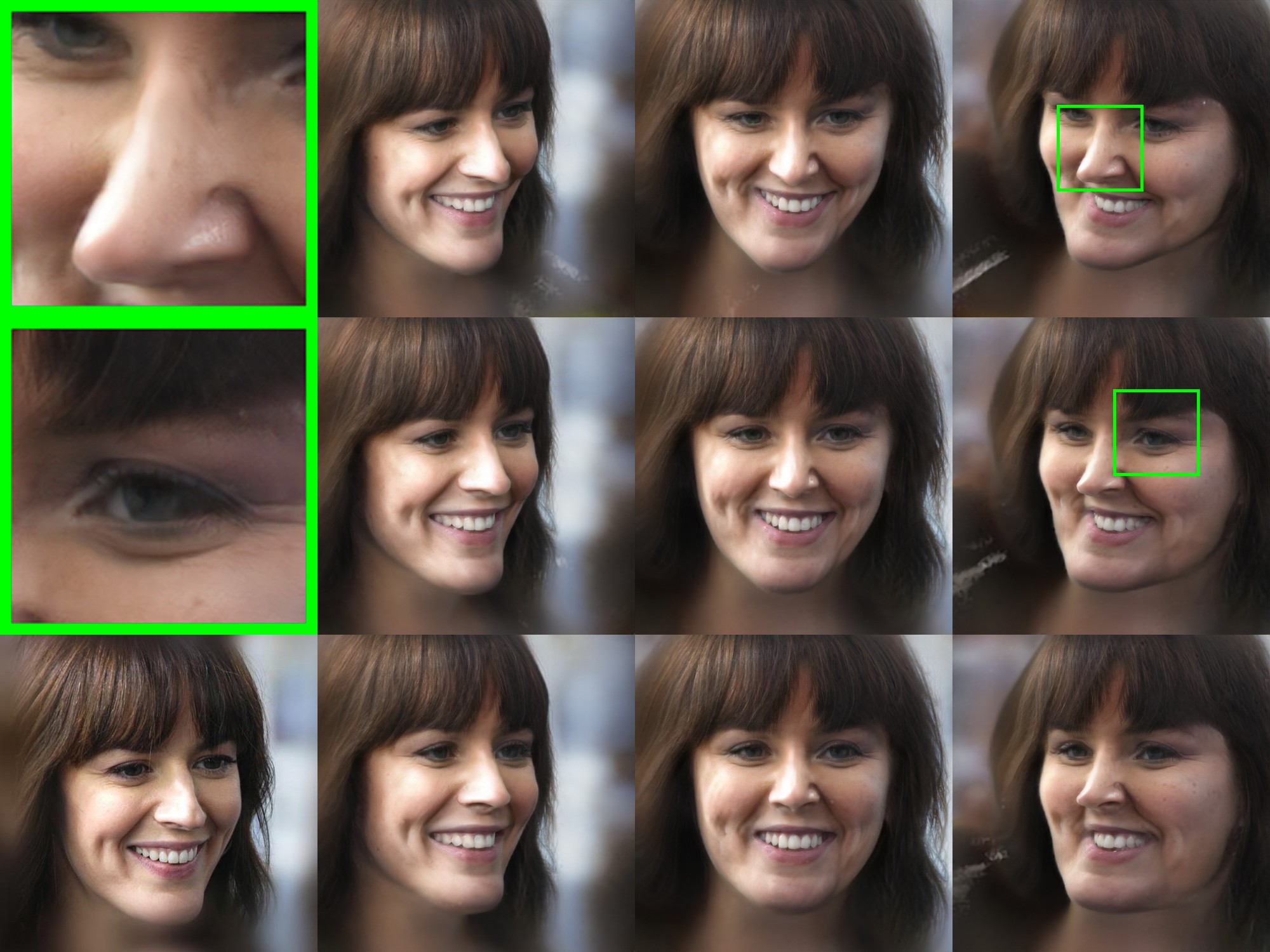}
\begin{tabularx}{0.4956\linewidth}{sb}
Input &  Novel Views 
\end{tabularx}
\begin{tabularx}{0.4956\linewidth}{sb}
Input &  Novel Views 
\end{tabularx}
\begin{tabularx}{\linewidth}{FF}
Ray Conditioning + PTI~\cite{roich2021pivotal} & EG3D~\cite{Chan2021} + PTI~\cite{roich2021pivotal}
\end{tabularx}\\
\vspace{15pt}

\includegraphics[width=0.495\linewidth]{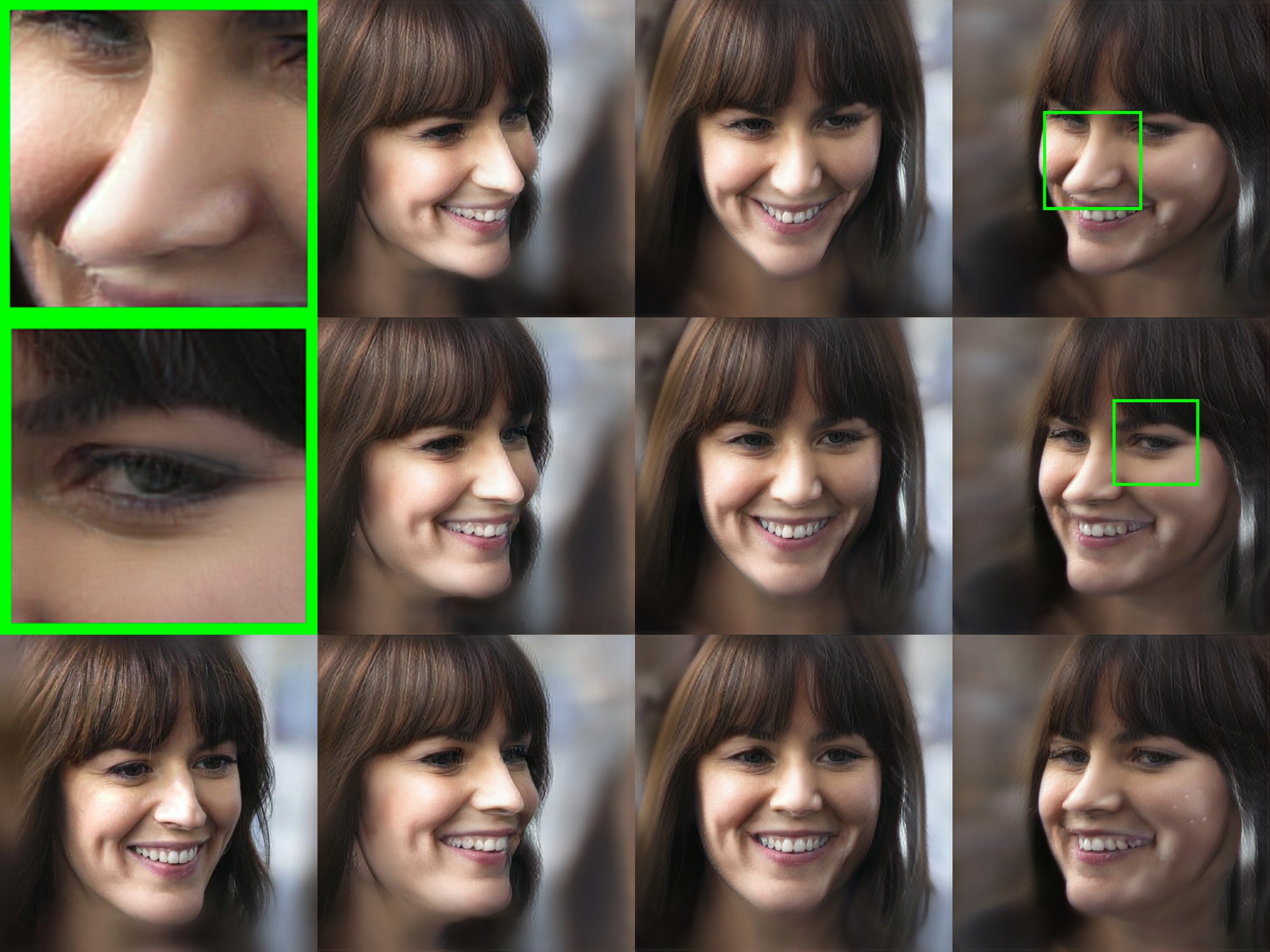}
\includegraphics[width=0.495\linewidth]{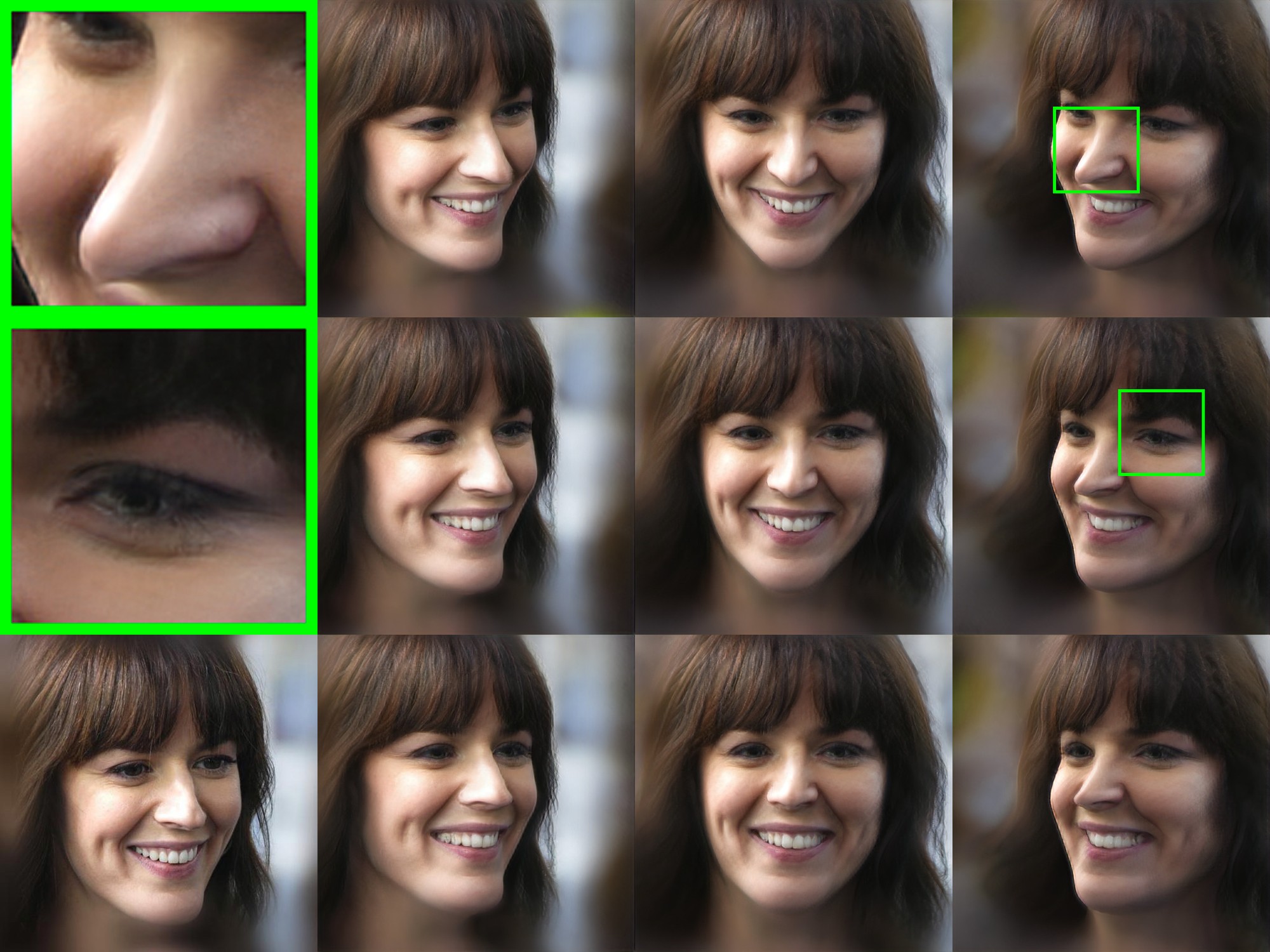}
\begin{tabularx}{0.4956\linewidth}{sb}
Input &  Novel Views 
\end{tabularx}
\begin{tabularx}{0.4956\linewidth}{sb}
Input &  Novel Views 
\end{tabularx}
\begin{tabularx}{\linewidth}{FF}
EG3D~\cite{Chan2021} + 3D GAN Inversion~\cite{ko20233d} & EG3D~\cite{Chan2021} + HFGI3D~\cite{xie2022HFGI3D}
\end{tabularx}\\
\vspace{15pt}

\caption{\textbf{Comparison to 3D-Aware GAN Inversion Methods.} Even dedicated 3D GAN inversion methods for EG3D~\cite{ko20233d,xie2022HFGI3D} can struggle to generate realistic novel views. It is especially challenging for 3D-aware GAN inversion methods when the input image is not front facing. It tends to distort the geometry of the input individual. For instance, when combined with EG3D, 3D GAN inversion~\cite{ko20233d} and PTI~\cite{roich2021pivotal} appear to widen the individual's face. PTI~\cite{roich2021pivotal}, 3D GAN Inversion~\cite{ko20233d}, and HFGI3D~\cite{xie2022HFGI3D} all appear to make the individual's nose more pointy.}
\label{fig:3d_inversion}
\end{figure*}

\begin{figure*}
\centering
\newcolumntype{b}{F}
\newcolumntype{s}{>{\hsize=.5\hsize}F}

\includegraphics[width=0.495\linewidth]{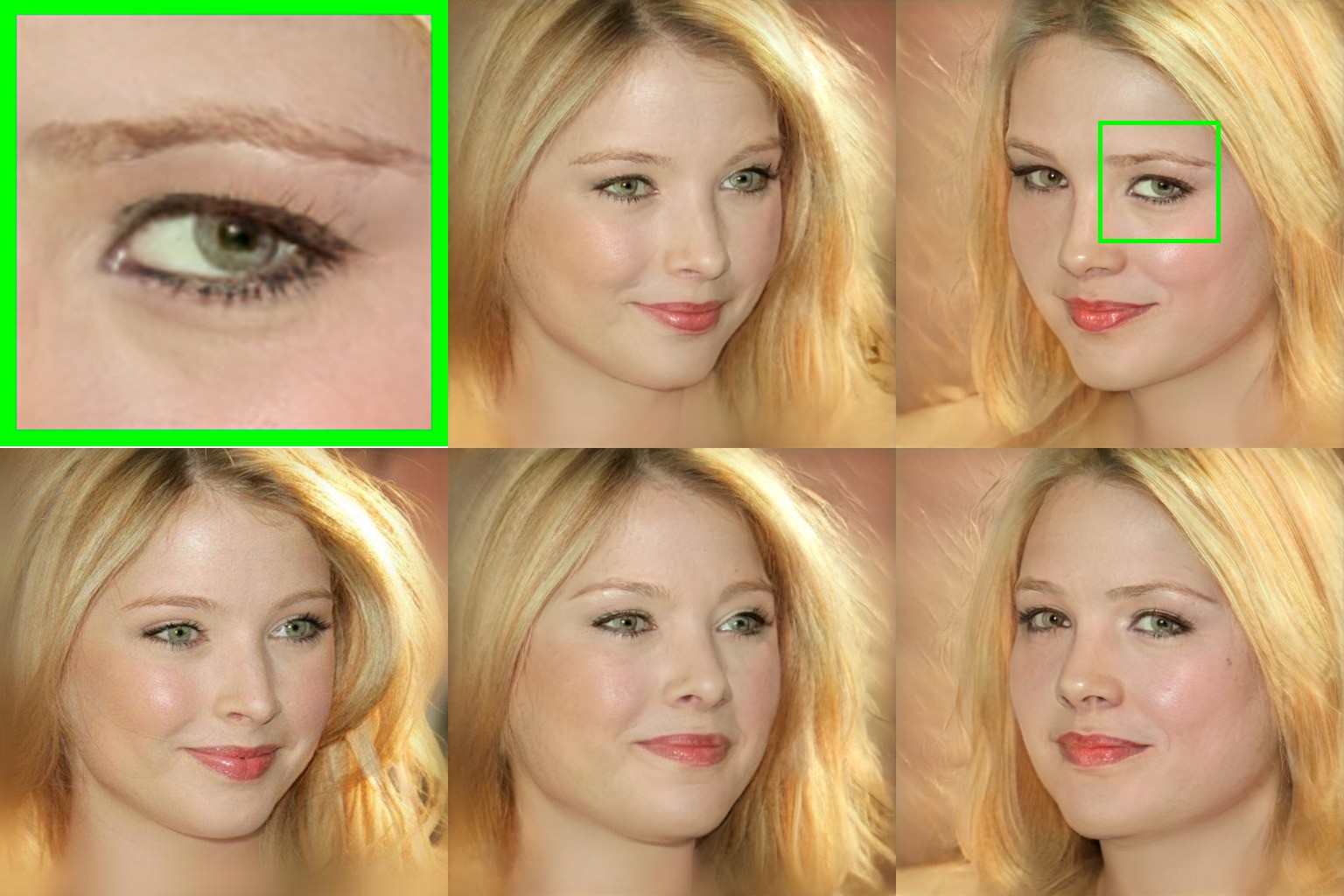}
\includegraphics[width=0.495\linewidth]{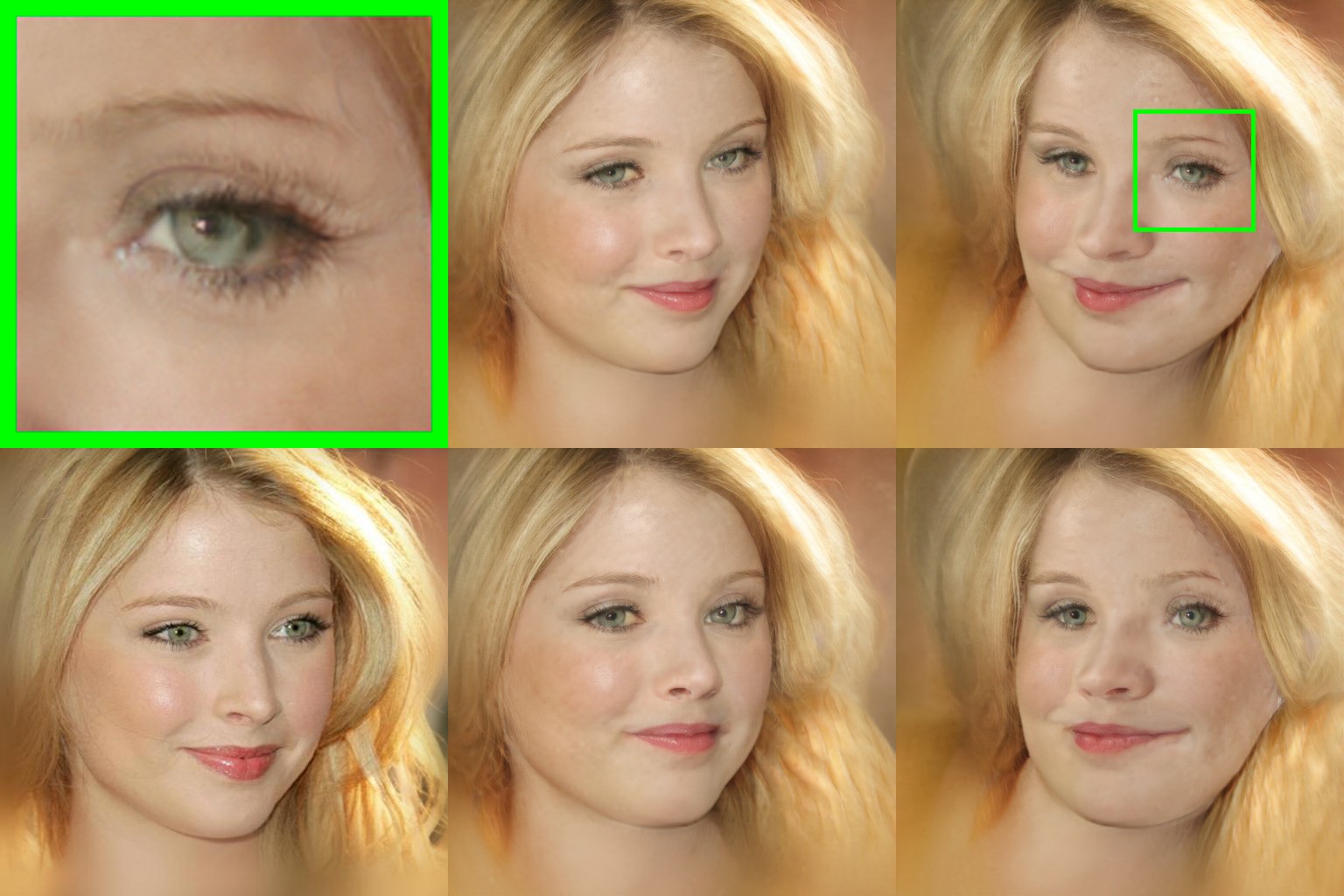}
\begin{tabularx}{0.4956\linewidth}{sb}
Input &  Novel Views 
\end{tabularx}
\begin{tabularx}{0.4956\linewidth}{sb}
Input &  Novel Views 
\end{tabularx}
\begin{tabularx}{\linewidth}{FF}
Ray Conditioning + PTI~\cite{roich2021pivotal} & EG3D~\cite{Chan2021} + PTI~\cite{roich2021pivotal}
\end{tabularx}\\
\vspace{15pt}

\includegraphics[width=0.495\linewidth]{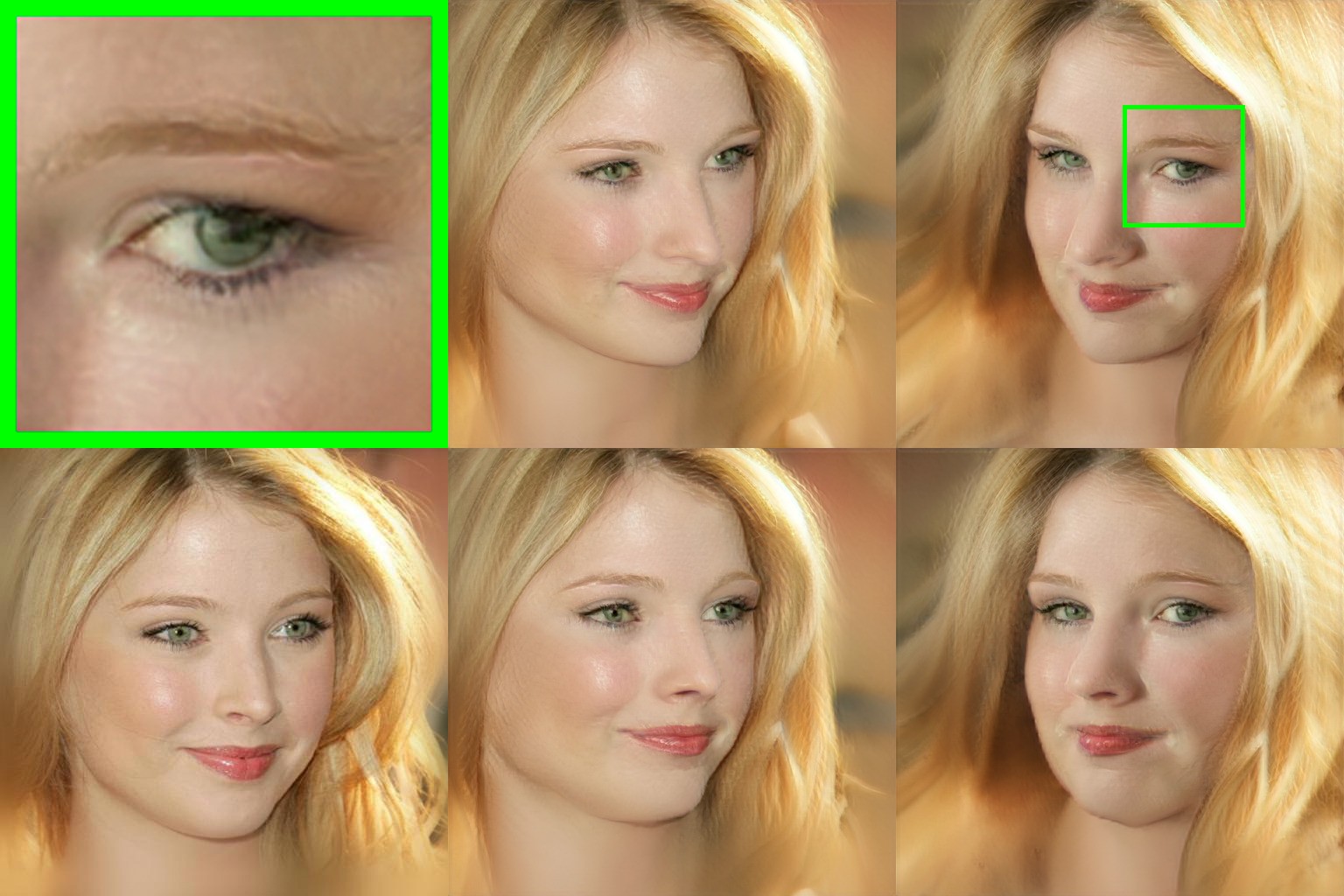}
\includegraphics[width=0.495\linewidth]{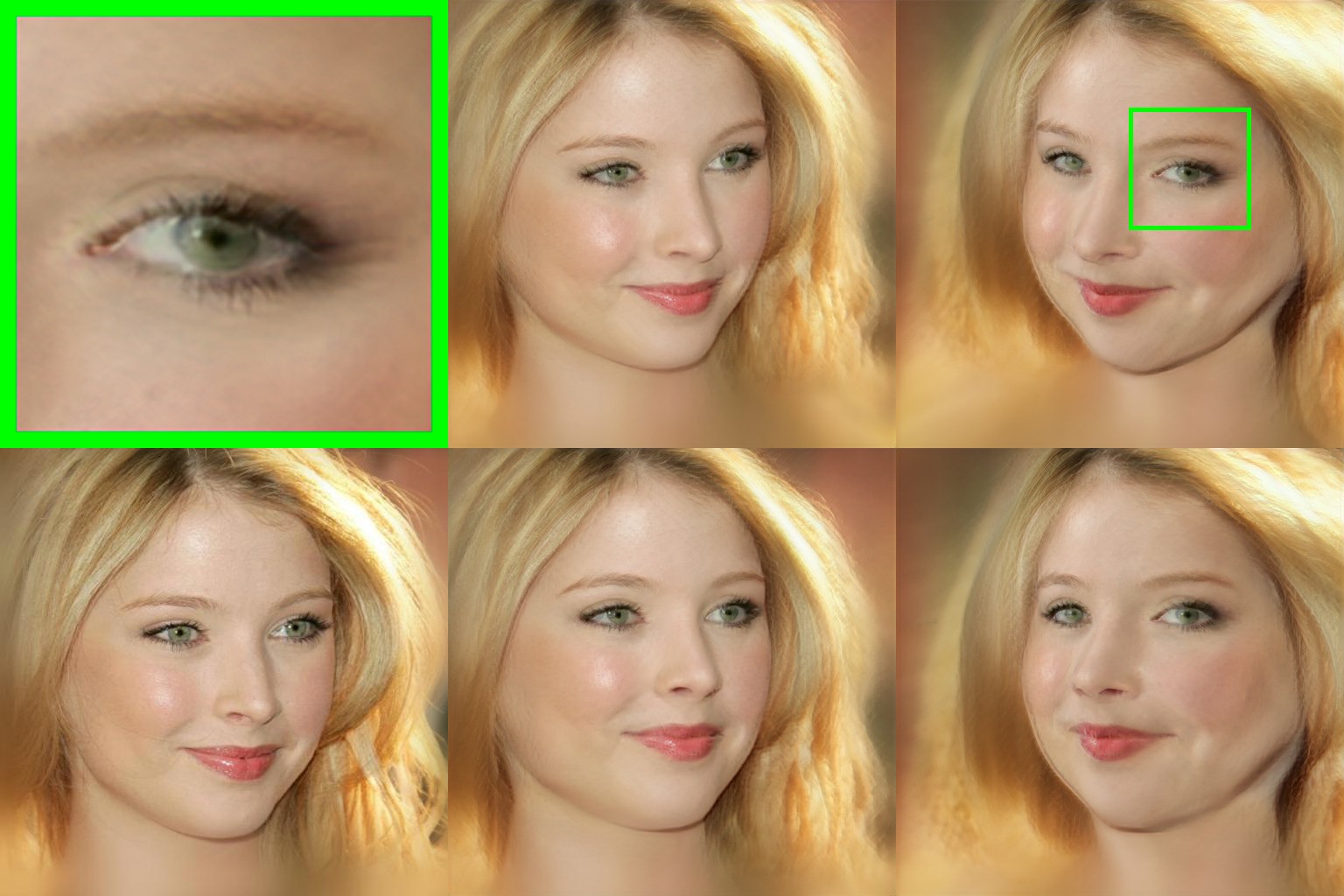}
\begin{tabularx}{0.4956\linewidth}{sb}
Input &  Novel Views 
\end{tabularx}
\begin{tabularx}{0.4956\linewidth}{sb}
Input &  Novel Views 
\end{tabularx}
\begin{tabularx}{\linewidth}{FF}
EG3D~\cite{Chan2021} + 3D GAN Inversion~\cite{ko20233d} & EG3D~\cite{Chan2021} + HFGI3D~\cite{xie2022HFGI3D}
\end{tabularx}\\
\vspace{15pt}

\caption{\textbf{Comparison to 3D-Aware GAN Inversion Methods.} Ray conditioning provides the best image quality compared to geometry-based GANs and inversion methods. The loss of image quality is noticeable. For 3D GAN Inversion~\cite{ko20233d} and HFGI3D~\cite{xie2022HFGI3D}, there are streaks across the face, hinting at spatial aliasing issues. 3D GAN Inversion and HFGI3D also cannot reconstruct the specularity of the eyes, which is very apparent in the ray conditioning example. }
\label{fig:3d_inversion_2}
\end{figure*}

\begin{figure*}
\centering
\newcolumntype{b}{F}
\newcolumntype{s}{>{\hsize=.3\hsize}F}
\includegraphics[width=0.49\linewidth]{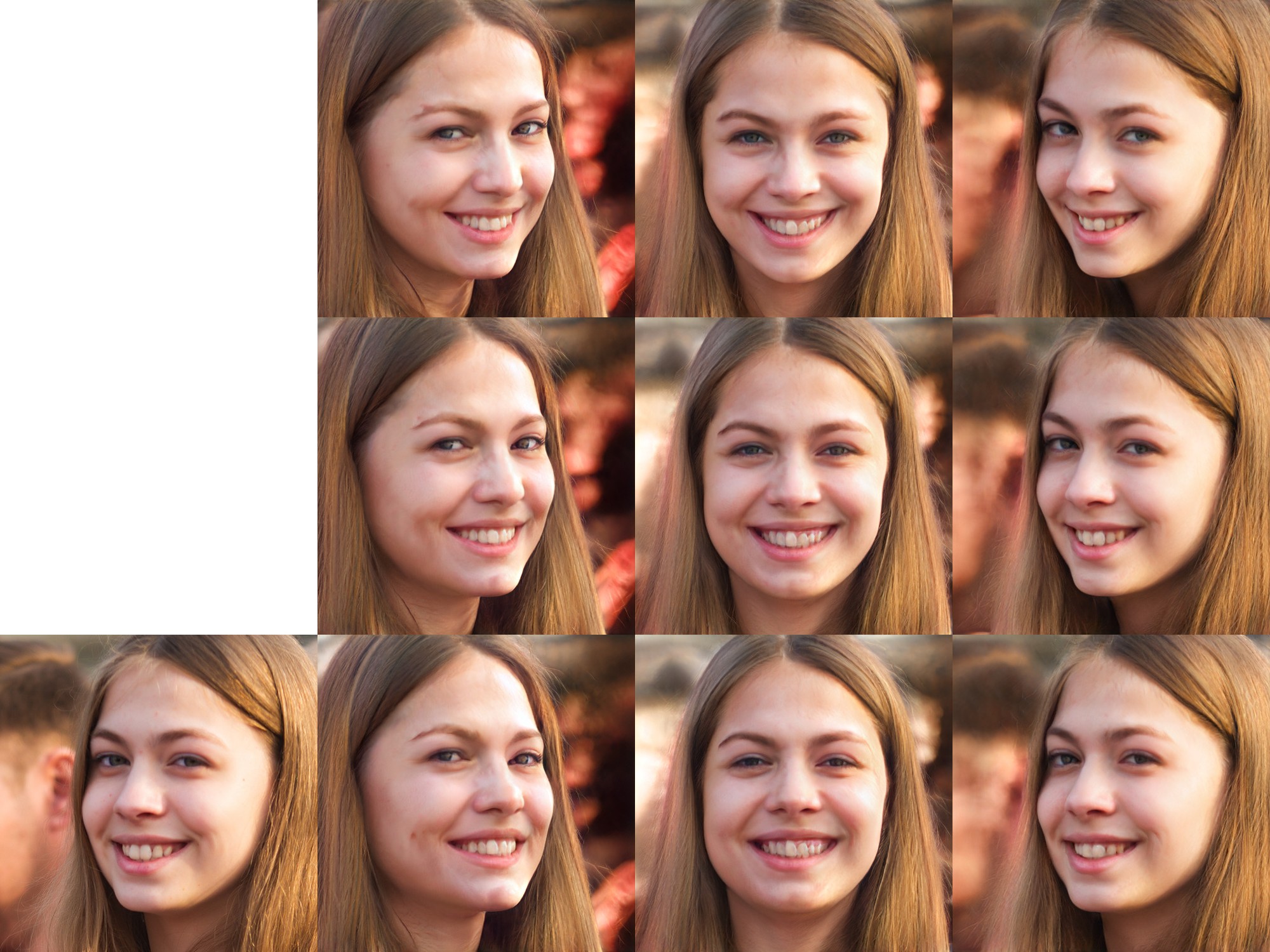}
\includegraphics[width=0.49\linewidth]{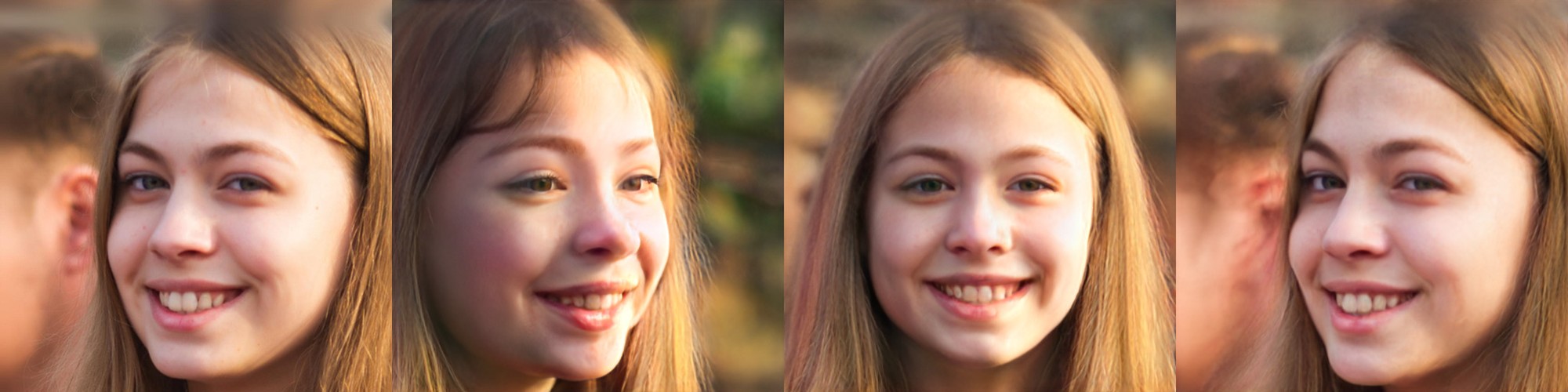}
\begin{tabularx}{0.496\linewidth}{sb}
Input &  Novel Views 
\end{tabularx}
\begin{tabularx}{0.496\linewidth}{sb}
Input &  Novel Views 
\end{tabularx}
\begin{tabularx}{\linewidth}{FF}
Ray Conditioning & InterfaceGAN~\cite{shen2020interfacegan}
\end{tabularx}

\caption{\textbf{Comparison to InterfaceGAN.} Latent space editing techniques such as InterfaceGAN~\cite{shen2020interfacegan} can generate binary changes to images to make them left facing or right facing. However, these models lack the same level of control that ray conditioning and geometry-based generative models do. We can achieve free viewpoint control with ray conditioning, while latent space editing is restricted to one dimension. InterfaceGAN also lacks the same amount of disentanglement as we do, causing identity shift.}
\label{fig:interfacegan}
\end{figure*}

In terms of image quality, ray conditioning is a large improvement over Light Field Networks (LFNs)~\cite{Sitzmann2021LFNs}.  We provide more results from LFNs on FFHQ in Figure~\ref{fig:lfns_faces}. As discussed in the main paper, LFNs have two main challenges. First, LFNs struggle to reconstruct high frequency details on a photo-realistic dataset such as FFHQ.  We also attempted to train LFNs with SIREN~\cite{sitzmann2019siren} activations instead of ReLU activations, but the model struggled to converge. Second, LFNs are unable to generate novel views when trained on a dataset with only one image per instance. Our work demonstrates that light field priors introduced in LFNs can be naturally extended from MLPs to more powerful CNN-based image synthesizers.

\section{Evaluation Details}
\label{app:eval}
\subsection{Datasets}
Similar to prior work~\cite{Chan2021,zhao-gmpi2022}, our method requires a dataset of images and estimated camera poses. We outline the datasets used below. 

\begin{figure*}
\centering

\includegraphics[width=0.49\linewidth]{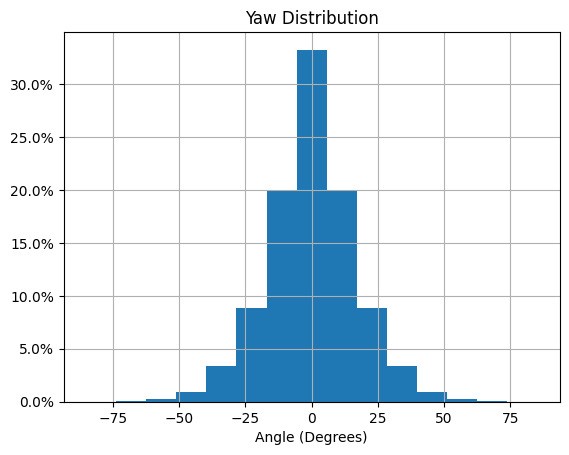}
\includegraphics[width=0.49\linewidth]{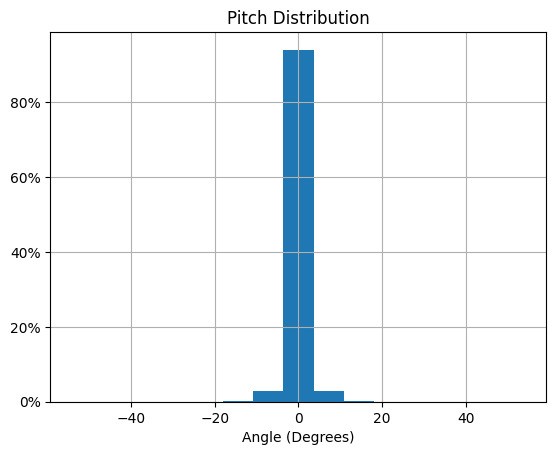}

\caption{\textbf{FFHQ Pose Distribution.} FFHQ consists mainly of front facing photographs. Its standard deviation for yaw (horizontal rotation) is $16^\circ$. Its standard deviation for pitch (vertical rotation) is $2^\circ$ degrees. As noted by EG3D~\cite{Chan2021}, this unbalanced distribution of subject pose is a challenge for all multi-view generative models trained on FFHQ. Proper data augmentation to reduce distribution bias is still an open and important problem.}
\label{fig:ffhq_angle_dist}
\end{figure*}

\begin{figure*}
  \centering
  \includegraphics[width=\textwidth]{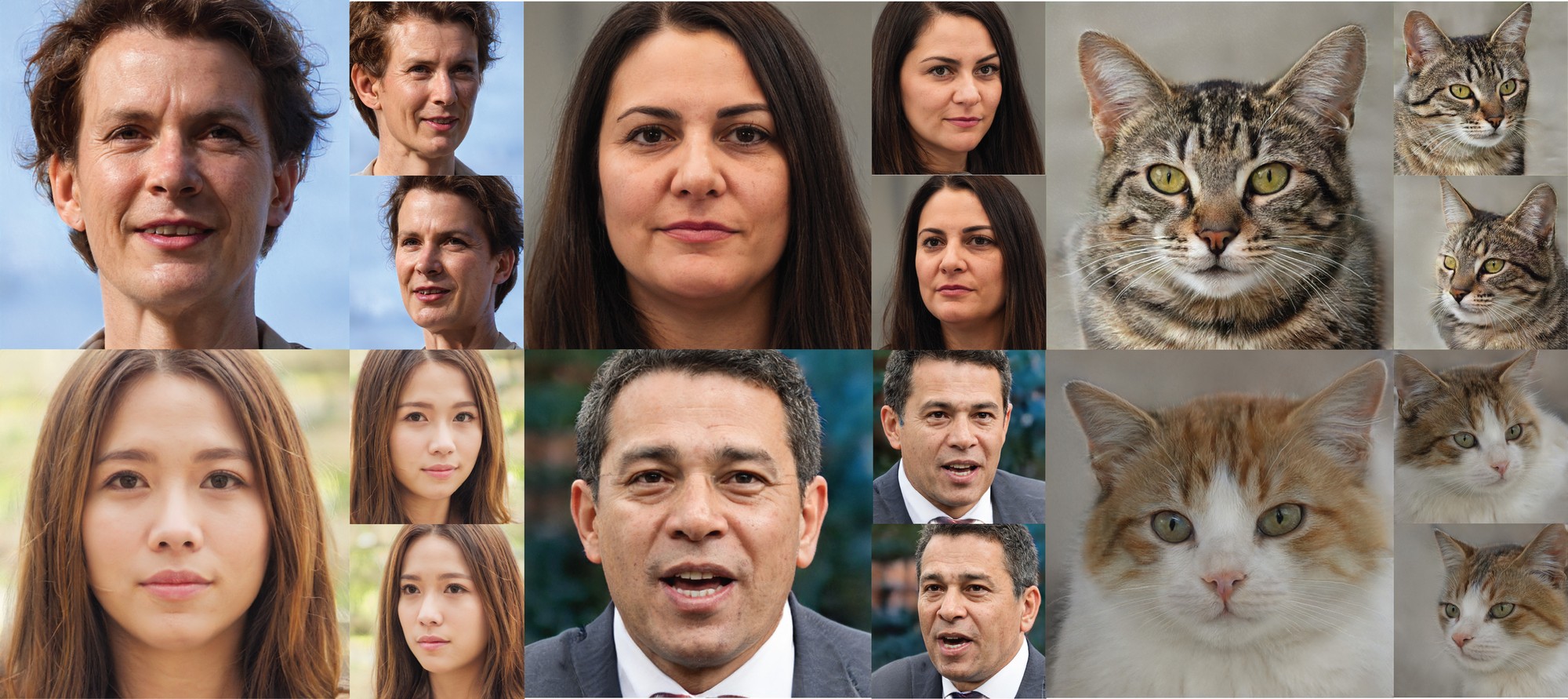}\\
  \vspace{20pt}
  \includegraphics[width=\textwidth]{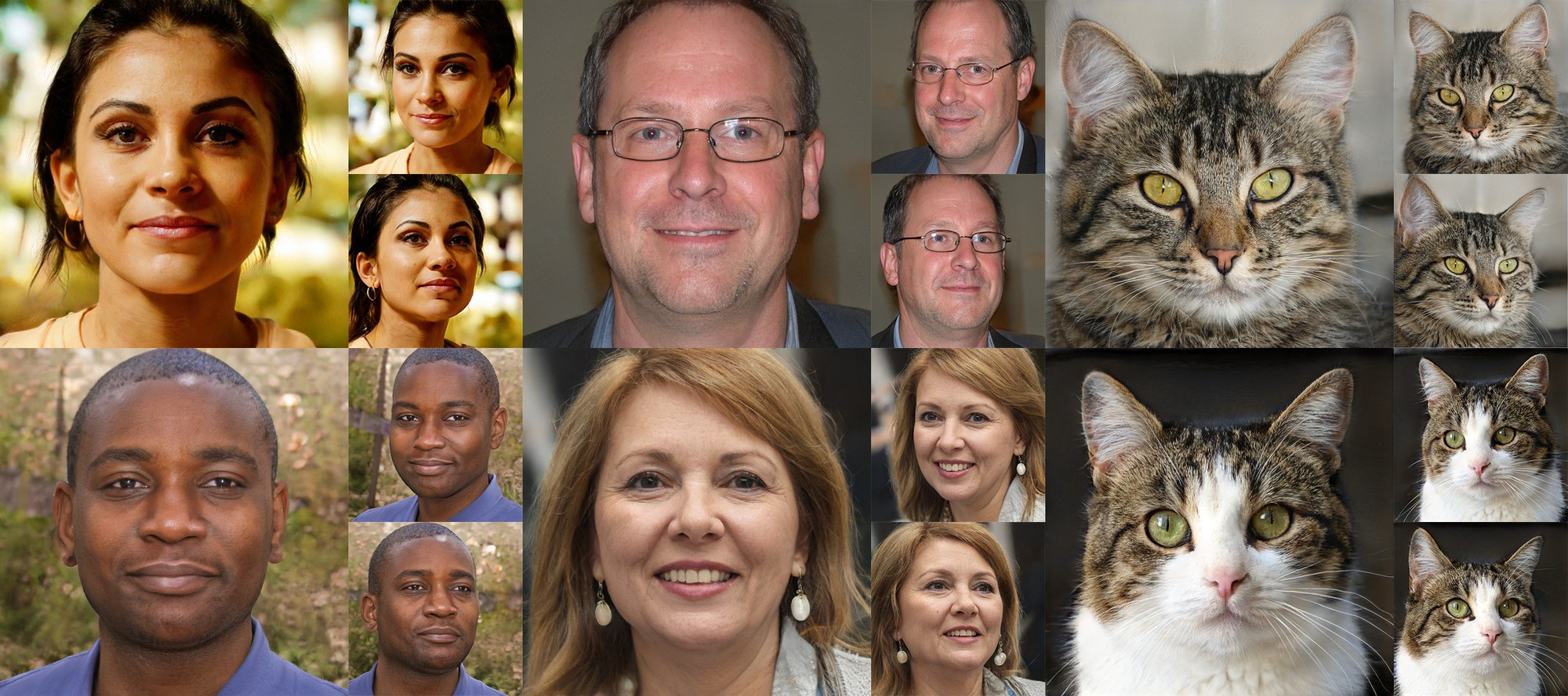}
  \caption{\textbf{Curated Latent Space Samples.} Given the 12 front facing images, we show that we can map them to a different viewpoint while maintaining the quality of the generated faces.}
  \label{fig:sample-results}
\end{figure*}

\begin{figure*}
\centering
\includegraphics[width=\linewidth]{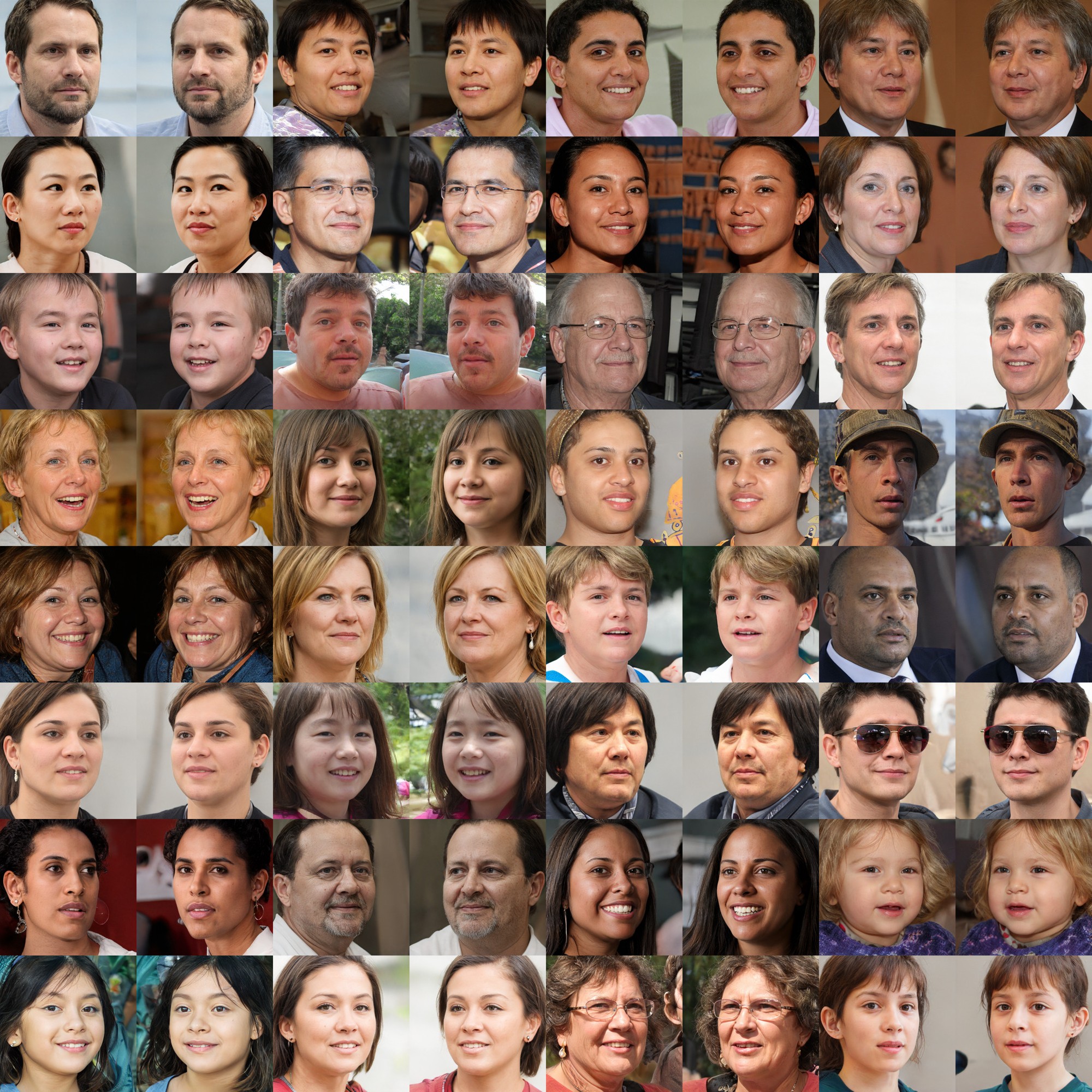}

\caption{\textbf{Uncurated Latent Samples.} We picture two views from latent vectors of seeds 0-31 to demonstrate the quality of our GAN. Even without a 3D representation, ray conditioning successfully generates images of the same individual from different view points. Results were generated with a truncation of $\psi = 0.7$.}
\label{fig:latent_samples}
\end{figure*}

\noindent\textbf{FFHQ Human Faces.} FFHQ~\cite{Karras2018ASG} is a dataset of $\sim70$k $1024\times1024$ images of front-facing faces. We use camera poses provided by EG3D, which are estimated by a deep face pose estimator~\cite{deng2019accurate}; camera poses are assumed to be distributed on a sphere, all facing a shared center. As previously reported by EG3D, this dataset contains bias which may affect the resulting generations. For instance, people in front-facing images are more likely to smile. People who appear to be lower than the camera tend to be children. In Figure~\ref{fig:ffhq_angle_dist}, we present the distribution of subject pose in terms of yaw (horizontal rotation), and pitch (vertical rotation). 


\noindent\textbf{AFHQv2 Cat Faces}. AFHQv2~\cite{choi2020starganv2} is a dataset consisting of many animal faces. We train our model on the cat subset using the camera poses provided by EG3D. This subset only consists of $\sim5$k $512\times512$ images, which is much smaller than FFHQ. Some pretraining is expected for good results. 

\noindent\textbf{SRN Cars}. The SRN Cars~\cite{sitzmann2019srns} training set is a collection of $\sim$2.5k ShapeNet~\cite{Shapenet2015} cars, each rendered from $250$ cameras distributed on a sphere at a resolution of $128 \times 128$.
Because it contains multiple images per object, it is commonly used for evaluating geometry-free view synthesis models.  We demonstrate our method's ability to generate $360^\circ$ light fields, and compare to the LFNs~\cite{Sitzmann2021LFNs} baseline on this dataset. Unlike for FFHQ or AFHQ, we start training from scratch.

\end{document}